\documentclass[11pt]{article}

\usepackage[preprint]{acl}
\usepackage{booktabs,xcolor,colortbl}
\usepackage{times,booktabs,multirow}
\usepackage{latexsym,amsmath,amssymb,amsthm}
\usepackage[T1]{fontenc}
\usepackage[utf8]{inputenc}
\usepackage{microtype}
\usepackage{inconsolata}
\usepackage{graphicx}
\usepackage[most]{tcolorbox}
\usepackage{dsfont}
\usepackage{mathtools}
\usepackage{amssymb}
\usepackage{subcaption}
\newtcolorbox{takeaway}[1][]{
  enhanced,
  colback=blue!3,
  colframe=blue!50!black,
  boxrule=0.4pt,
  arc=2pt,
  left=8pt, right=8pt, top=6pt, bottom=6pt,
  fonttitle=\bfseries\small,
  title=Key Takeaway,
  coltitle=blue!50!black,
  attach boxed title to top left={xshift=10pt, yshift=-7pt},
  boxed title style={colback=white, colframe=white, boxrule=0pt, left=2pt, right=2pt, top=0pt, bottom=0pt},
  #1
}
\usepackage[utf8]{inputenc}
\usepackage{xcolor,mdframed}
  \definecolor{xmlcol}{RGB}{173,214,241}        
  \definecolor{jsoncol}{RGB}{180,230,180}       
  \definecolor{loosecol}{RGB}{255,235,160}      
  \definecolor{freecol}{RGB}{250,190,190}       
  \definecolor{logitcol}{RGB}{210,210,210}      

  \mdfdefinestyle{xmlbox}{backgroundcolor=xmlcol!40,linecolor=xmlcol,linewidth=0.8pt,
    innerleftmargin=6pt,innerrightmargin=6pt,innertopmargin=4pt,innerbottommargin=4pt,
    skipabove=4pt,skipbelow=4pt}
  \mdfdefinestyle{jsonbox}{backgroundcolor=jsoncol!40,linecolor=jsoncol,linewidth=0.8pt,
    innerleftmargin=6pt,innerrightmargin=6pt,innertopmargin=4pt,innerbottommargin=4pt,
    skipabove=4pt,skipbelow=4pt}
  \mdfdefinestyle{loosebox}{backgroundcolor=loosecol!40,linecolor=loosecol,linewidth=0.8pt,
    innerleftmargin=6pt,innerrightmargin=6pt,innertopmargin=4pt,innerbottommargin=4pt,
    skipabove=4pt,skipbelow=4pt}
  \mdfdefinestyle{freebox}{backgroundcolor=freecol!40,linecolor=freecol,linewidth=0.8pt,
    innerleftmargin=6pt,innerrightmargin=6pt,innertopmargin=4pt,innerbottommargin=4pt,
    skipabove=4pt,skipbelow=4pt}
  \mdfdefinestyle{logitbox}{backgroundcolor=logitcol!40,linecolor=logitcol,linewidth=0.8pt,
    innerleftmargin=6pt,innerrightmargin=6pt,innertopmargin=4pt,innerbottommargin=4pt,
    skipabove=4pt,skipbelow=4pt}

\newcommand{\Ta}{\cellcolor{green!40}}   
\newcommand{\Tb}{\cellcolor{green!18}}   
\newcommand{\Tc}{\cellcolor{yellow!25}}  
\newcommand{\Td}{\cellcolor{orange!25}}  
\newcommand{\Te}{\cellcolor{red!25}}     
\newcommand{\sg}{\cellcolor{black!4}}    
\newcommand{\sm}{\cellcolor{black!9}}    
\newcommand{\satb}{\cellcolor{black!16}} 
\title{All Verdicts are Not Equal: Rethinking LLM Judge Reliability}

\author{Vineet Kumar\\
  PayPal Artificial Intelligence\\
  PayPal, Bengaluru, India\\
  \texttt{vkumar32@paypal.com} \\\And
  Darshita Rathore \\
  PayPal Artificial Intelligence\\
  PayPal, Bengaluru, India\\
  \texttt{drathore@paypal.com} \\\And
    Anindya Moitra\\
  PayPal Artificial Intelligence\\
  PayPal, Bengaluru, India\\
  \texttt{amoitra@paypal.com}\\
  }

\begin{document}
\maketitle

\begin{abstract}
LLM-as-a-Judge is the standard paradigm for NLP evaluation, yet its systemic reliability remains poorly understood despite being widely treated as a deterministic ground truth. We present a comprehensive reliability audit, stress-testing six frontier models across four benchmarks, five prompt formats, two presentation orders, three sampling temperatures, and ten repetitions per condition. Our empirical analysis reveals severe vulnerabilities: verdicts change across identical replications at temperature zero, position-order swaps flip the majority of verdicts on challenging tasks, and the most deterministic judge achieves perfect consistency by trivially repeating incorrect verdicts, agreeing with ground truth only 51\% of the time. To formalize these multi-faceted failure modes, we introduce the trustworthy verdict rate ($T$), a unified metric capturing the joint probability that an evaluation is reproducible, order-invariant, and accurate. Using $T$, we derive a theoretical upper bound on accuracy imposed by position bias and show that reliability is item-specific rather than model-level. Finally, we demonstrate that shifting from pairwise win-rate to holistic rubric scoring improves trustworthiness more than any single-format prompting intervention, offering an actionable framework for robust NLP evaluation.
\end{abstract}  
\section{Introduction}
\label{sec:intro}
LLM-as-a-Judge has become the dominant evaluation paradigm in
modern NLP. It powers RL reward signals \cite{wen2026reinforcement}, model-comparison
 \cite{dubois2023alpacafarm}, automated benchmark scoring, and an increasing
fraction of the evaluation pipelines on which downstream model
selection now depends \citep{zheng2023judging,
tan2025judgebench}. Across all of
these settings, a judge's output is treated as a stand-in for
ground truth: the verdict is what gets logged, aggregated, and
optimized against. 

This treatment carries an implicit assumption that the
verdict is a stable function of the inputs. That re-querying the
same judge with the same pair yields the same answer, and that
swapping which response is shown first does not change the
preferred one. Both assumptions fail in practice. In our audit,
7--17\% of verdicts change across identical replications, and up to 98\% of verdicts flip when the
presentation order of the two candidates is swapped. The most
striking case in our data is a judge that returns the same
verdict on 100\% of repeated calls maximally deterministic by
every single-shot diagnostic in common use yet whose verdicts
agree with ground truth only 51\% of the time, barely above
chance. Determinism, in other words, is not reliability.

We address this with two complementary contributions. First, a
\textbf{cross-cutting empirical audit}: we stress-test six
frontier judges across four
benchmarks spanning subjective and objective preference,
standard and adversarial difficulty, at five prompt formats,
two presentation orders, three sampling temperatures, and ten
repetitions per condition, one of the largest reliability-focused
audits of LLM-as-Judge systems we are aware of. The design is
deliberately cross-cutting: each axis of variation is held
fixed in turn so that its individual contribution to verdict
instability can be measured rather than averaged out.

Second, a \textbf{formal framework} that consolidates the
audit's findings into operationally meaningful quantities. We
argue that reliability is irreducibly multidimensional and
introduce the \emph{trustworthy verdict rate} as the joint
probability that a verdict is reproducible, order-invariant,
and correct. The framework yields a hard accuracy ceiling
imposed by position bias that no improvement in model
capability can exceed, decomposes reliability loss into a noise
component and a position-bias component that respond to
different interventions, and produces a per-cell scalar that
reorders judge comparisons relative to accuracy whenever judges
differ on the underlying axes. We further
compare three evaluation paradigms (direct pairwise, FLASK
rubric decomposition, and G-Eval holistic scoring) and find that
paradigm choice can shift trustworthiness further than any
single-format intervention.

Our findings show that accuracy alone is insufficient to characterize LLM judges, single-order protocols mask critical position bias, and common single-shot consistency heuristics fail to catch major systemic failure modes.

The remainder of this paper is structured as follows: we survey related literature in \S\ref{sec:related_works}, establish our formal multidimensional reliability framework in \S\ref{sec:formalization}, outline our extensive empirical audit setup in \S\ref{sec:expt}, analyze our results and findings in \S\ref{sec:findings}, and conclude with actionable recommendations in \S\ref{sec:conclusion}.

\section{Related Works}
\label{sec:related_works}

LLMs are used across domains codegen \cite{genicious}, conversational analytics \cite{rathore2026conversationalqueryenginemixedmodality}, agentic evaluation \cite{agarwal2026agentworld} and LLM-based evaluation has become a standard alternative to human annotation/evaluation. With MT-Bench \cite{zheng2023judging} showing that GPT-4 judgments can achieve high agreement with human preferences. The paradigm has since been extended to RLHF simulation in AlpacaFarm \cite{dubois2023alpacafarm}, objective ground-truth evaluation in JudgeBench \cite{tan2025judgebench}, and adversarial robustness in LLMBar \cite{zeng2024evaluating}. Recent surveys organize the field around what to judge, how to judge, and how to benchmark \cite{li-etal-2025-generation}. Most of this work reports agreement or accuracy; we instead ask whether a judge's verdict remains reliable under repeated sampling, position swaps, and format changes.

A growing literature shows that LLM judges are sensitive to response order and presentation effects \cite{wang2023large, shi2024judging}, can be shifted by adversarial phrases \cite{raina-etal-2024-llm}, and may be overconfident on ambiguous examples despite modest accuracy \cite{lu-etal-2025-llm}. Complementary work finds low intra-rater reliability and overconfident inconsistency in LLM judges \cite{haldar-hockenmaier-2025-rating, janiak-etal-2025-illusion}. We build on these observations by formalizing position sensitivity as a flip rate $\varphi$, deriving the induced accuracy ceiling $1-\varphi/2$, and combining it with stochastic consistency $C$ into the Trustworthy Verdict Rate $T$.

Our treatment also connects to uncertainty estimation and structured evaluation. Selective prediction studies how classifiers trade coverage for accuracy by abstaining on uncertain inputs \cite{elyaniv2010foundations}, and recent work adapts this idea to LLM judges using conformal prediction with held-out calibration data \cite{chirkova-etal-2026-llm}. Other studies show that evaluation format affects judge behavior: averaging scores can outperform majority vote, chain-of-thought can hurt some judges, and checklist-style rubrics can improve agreement \cite{sheng-etal-2025-analyzing, lee-etal-2025-checkeval}. Our results connect these strands by showing that repeated-query agreement provides a calibration-free confidence signal, while G-Eval-style single-prompt multi-criterion scoring improves $T$ more reliably than decomposed rubric voting.

\section{Formalization}
\label{sec:formalization}

\subsection{Setup}
\label{sec:setup}

We evaluate a judge $J$ on a corpus of preference pairs. Each instance comprises two candidate responses $(x, y)$ together with a ground-truth label indicating which is preferred. The judge returns a verdict $V \in \{x, y\}$ for any presentation of the pair.

Reliability is studied along three axes:
\begin{itemize}
    \item \textbf{Stochastic instability}: the verdict varies across independent runs of the same prompt.
    \item \textbf{Position bias}: the verdict varies with the order in which the two candidates are presented.
    \item \textbf{Format sensitivity}: the verdict statistics vary with the prompt template (e.g., direct vs.\ CoT, JSON vs.\ XML).
\end{itemize}

The first two axes act on a single elicitation and enter the trustworthy rate $T$ defined below. The third axis is treated as a \emph{design factor}: we fix a set $\mathcal{F}$ of prompt templates and compute every quantity in \S\ref{sec:indicators}--\S\ref{sec:trustworthy} separately for each $f \in \mathcal{F}$. Format sensitivity is then characterized in \S\ref{sec:format} by the dispersion of those per-format quantities for a fixed (judge, dataset).

\subsection{Per-Pair Indicators and Population Rates}
\label{sec:indicators}

For each pair, holding the format $f$ and dataset fixed, we define three Bernoulli indicators:
\begin{align*}
    N &= \mathds{1}[\small{\text{verdict reproducible across independent runs}}] \\
    P &= \mathds{1}[\small{\text{verdict invariant under candidate-order swap}}] \\
    R &= \mathds{1}[\small{\text{verdict matches ground truth}}]
\end{align*}

The corresponding population rates are directly observable from repeated elicitations:
\begin{align*}
    C &\coloneqq \Pr(N = 1) && \text{(consistency rate)} \\
    1 - \varphi &\coloneqq \Pr(P = 1) && \text{($\varphi$ is the position flip rate)} \\
    A &\coloneqq \Pr(R = 1) && \text{(overall accuracy)}
\end{align*}

We partition pairs by $P$ into \emph{decisive} ($P = 1$) and \emph{undecisive} ($P = 0$) subsets, letting $A_{\text{dec}} \coloneqq \Pr(R = 1 \mid P = 1)$ denote the decisive accuracy and $q \coloneqq A_{\text{undec}} = \Pr(R = 1 \mid P = 0)$ denote the accuracy on undecisive pairs.

\subsection{Assumptions}
\label{sec:assumptions}

\paragraph{(A1) Undecisive-pair accuracy.} In general evaluation protocols involving ties, abstentions, non-binary labels, or unequal order weighting, $q \coloneqq \Pr(R = 1 \mid P = 0)$ captures baseline correctness on undecisive pairs. Under a forced-binary, equally weighted preference setup, any pair where the verdict flips under candidate swap yields contradictory verdicts across orderings, exactly one of which matches the ground truth. Hence, in our setup, $q = \tfrac{1}{2}$.\footnote{The assumption $q = \tfrac{1}{2}$ can break under explicit tie-breaking options or ``neither'' verdicts; however, in our core experiments, the judge produces a forced binary verdict.}

\paragraph{(A2) Axis independence on a pair.} Stochastic reproducibility is uninformative about position-decisiveness, and the correctness rate on consistent, decisive pairs equals the decisive accuracy:
\[
    \Pr(P = 1 \mid N = 1) = \Pr(P = 1)
\]
\[
    \Pr(R = 1 \mid N = 1, P = 1) = A_{\text{dec}}.
\]

\subsection{A Tight Upper Bound on Observable Accuracy}
\label{sec:ceiling}

Decomposing accuracy over the decisive partition,
\begin{equation}
    A = (1 - \varphi)\, A_{\text{dec}} + \varphi q,
    \label{eq:accuracy-decomp}
\end{equation}
so that
\begin{equation}
    A_{\text{dec}} = \frac{A - \varphi q}{1 - \varphi}.
    \label{eq:adec}
\end{equation}
Since $A_{\text{dec}} \leq 1$, we obtain the generalized \textbf{capability ceiling}
\begin{equation}
    \boxed{\;A \;\leq\; 1 - \varphi(1 - q)\;}
    \label{eq:ceiling}
\end{equation}
For our forced-binary setup ($q = \tfrac{1}{2}$), Eq.~\eqref{eq:ceiling} simplifies to \color{blue}$A \leq 1 - \varphi/2$\normalcolor.

The bound is tight and attained iff the judge is perfect on every decisive pair ($A_{\text{dec}} = 1$), and \emph{capability-free}: no improvement in the judge's underlying judgment can lift observable accuracy above $1 - \varphi(1 - q)$ when the position flip rate is $\varphi$. We call the residual $\bigl(1 - \varphi(1 - q)\bigr) - A$ the \textbf{capability gap}, the portion of error orthogonal to position bias.

\subsection{The Trustworthy Rate}
\label{sec:trustworthy}

A verdict is \textbf{trustworthy} when it is simultaneously reproducible, robust to candidate-order swap, and correct. The rate at which the judge produces such verdicts is
\[
    T \coloneqq \Pr(N = 1 \,\wedge\, P = 1 \,\wedge\, R = 1).
\]
Factoring by the chain rule and applying (A2)\footnote{In practice, this mutual independence and Assumption 2  (A2) may not hold. See Appendix \ref{app:a2_clarification} for more details. There, we empirically show that even when A2 is violated, $T$ continues to behave as a conservative lower bound.},
\[
\begin{aligned}
T &= \Pr(N) \cdot \Pr(P \mid N) \cdot \Pr(R \mid N, P) \\
  &= C \cdot (1 - \varphi) \cdot A_{\text{dec}}.
\end{aligned}
\]
Substituting~\eqref{eq:adec} yields the generalized closed form
\begin{equation}
    \boxed{\;T \;=\; C \cdot \bigl(A - \varphi q\bigr)\;}
    \label{eq:trustworthy}
\end{equation}
which simplifies to \color{blue} $T = C \cdot (A - \varphi/2)$ \normalcolor when $q = \tfrac{1}{2}$.

\paragraph{Interpretation.} Equation~\eqref{eq:trustworthy} factors $T$ into two observable terms:
\begin{itemize}
    \item $C$: \emph{noise survival}: the fraction of verdicts stable to re-sampling.
    \item $(A - \varphi q)$: \emph{capability net of undecisive-baseline slack}: the portion of accuracy not borrowed from the baseline $q$ on undecisive pairs.
\end{itemize}
Two judges with identical overall accuracy $A$ can therefore differ substantially in $T$ when their consistency or position flip rates differ. $T$ is the operationally meaningful quantity: the rate at which a judge produces verdicts that are reproducible, robust to order, and correct.

\subsection{Format as a Design Factor}
\label{sec:format}

We compute the tuple $(C, \varphi, A, T)$ separately for every (judge, format, dataset) triple in our experimental grid; each row of Table~\ref{tab:consistency_full} reports such a tuple. Format sensitivity is then summarized by within-cell dispersion of $T$ across $f \in \mathcal{F}$, holding judge and dataset fixed:
\begin{equation}
    \Delta T_{\text{fmt}}(J, D) \coloneqq \max_{f \in \mathcal{F}} T(J, f, D) \;-\; \min_{f \in \mathcal{F}} T(J, f, D).
    \label{eq:format-dispersion}
\end{equation}

A judge with small $\Delta T_{\text{fmt}}$ is format-robust; one with large $\Delta T_{\text{fmt}}$ has trustworthy verdicts that depend on a prompt-engineering choice. Reporting $T$ at the cell level lets readers inspect both \emph{level} (which (judge, format) combination is most trustworthy on a given benchmark) and \emph{robustness} (how much that level moves with format), without conflating the two into a single number.

\section{Experimental Design}

\label{sec:expt}

\subsection{Datasets}
\label{sec:setup:data}

We evaluate on four benchmarks spanning a 2$\times$2 design across subjective vs.\ objective and standard vs.\ adversarial settings. MT-Bench \cite{zheng2023judging} and AlpacaFarm \cite{dubois2023alpacafarm} represent standard subjective evaluation benchmarks, focusing respectively on crowd-sourced human preferences for open-ended responses and instruction-following preferences. For objective evaluation, we use JudgeBench \cite{tan2025judgebench}, which provides ground-truth labels derived from MMLU-Pro, LiveCodeBench, and LiveBench across 17 domains, and LLMBar \cite{zeng2024evaluating}, an adversarial benchmark containing rule-based response pairs specifically designed to test judge robustness.

\subsection{Judges}
\label{sec:setup:judges}

We evaluate 6 models from 4 providers: \textbf{GPT-5.2} and \textbf{GPT-4o} (OpenAI), \textbf{Gemini-2.5-Pro} and \textbf{Gemini-2.0-Flash} (Google), \textbf{Claude-Sonnet-4.6} (Anthropic), and \textbf{Llama-3.3-70B} \cite{llama31}.
This spans frontier closed-weight models and a strong open-weight model, with 2 reasoning-specialized models (GPT-5.2, Gemini-2.5-Pro) and 4 standard models.

\subsection{Prompt Formats}
\label{sec:setup:formats}

All formats share a common system prompt and user message template; only the output format instruction varies:

\begin{itemize}
  \item \textbf{Direct}: Single-token verdict (A or B). No rationale.
  \item \textbf{Verdict-first} (JSON / XML): Structured output with verdict \emph{before} rationale the ``decide then justify'' pattern.
  \item \textbf{CoT} (JSON / XML): Rationale \emph{before} verdict the ``think then decide'' pattern.
\end{itemize}

We focus on JSON and XML because they are the dominant structured-generation formats in modern LLM APIs and production pipelines, enabling us to isolate whether formatting conventions alone influence judge behavior. This creates a 2$\times$2 sub-design within structured prompting: \{JSON, XML\} $\times$ \{verdict-first, CoT\}.

Each pair is presented in both AB and BA order.
Every (pair, judge, format, order) combination is repeated 10 times.
We run at three sampling temperatures: $T_s{=}0$ (all 6 judges), $T_s{=}0.5$, and $T_s{=}1.0$ (4 judges; reasoning models excluded as public API does not support the temperature parameter). Detailed breakdowns of models utilized, and public API cost estimates are provided in Appendix~\ref{app:cost}.



\section{Results \& Empirical Findings}
\label{sec:findings}

Table~\ref{tab:consistency_full} reports stochastic consistency $C$, position flip rate $\varphi$, accuracy $A$, and Trustworthy Verdict Rate $T$ across six judges, five prompt formats, and four datasets. We organize the findings around the three empirical predictions of the formalization in \S\ref{sec:formalization}: position bias imposes a hard accuracy ceiling, $T$ separates judges that accuracy conflates, and stochastic consistency alone is insufficient for reliability. We then analyze the sources of variation in $T$ and evaluate whether rubric-based paradigms can mitigate the same failure modes.

\begin{table*}[h!]
\centering

\resizebox{\textwidth}{!}{%
\begin{tabular}{ll cccc cccc cccc cccc}
\toprule
\textbf{Judge} & \textbf{Format} & \multicolumn{4}{c}{\textbf{MT-Bench}} & \multicolumn{4}{c}{\textbf{AlpacaFarm}} & \multicolumn{4}{c}{\textbf{JudgeBench}} & \multicolumn{4}{c}{\textbf{LLMBar}} \\
\cmidrule(lr){3-6}\cmidrule(lr){7-10}\cmidrule(lr){11-14}\cmidrule(lr){15-18}
 &  & $C$ & $\varphi$ & $A$ & $T$ & $C$ & $\varphi$ & $A$ & $T$ & $C$ & $\varphi$ & $A$ & $T$ & $C$ & $\varphi$ & $A$ & $T$ \\
\midrule
\multirow{5}{*}{\textbf{GPT-5.2}}
 & Direct            & \sg 89.3 & \sg 5.1  & \sm 64.8 & \Tc 55.6 & \sm 82.7 & \sg 9.2  & \satb 49.5 & \Te 37.1 & \sg 86.0 & \sm 11.0 & \sm 64.5 & \Tc 50.7 & \sg 89.0 & \sg 9.0  & \sg 75.5 & \Tb 63.2 \\
 & VF -- JSON        & \sm 86.2 & \sg 7.1  & \sm 64.3 & \Tc 52.4 & \sm 82.1 & \sg 6.1  & \sm 52.0 & \Td 40.2 & \sm 81.0 & \sm 10.0 & \sm 65.0 & \Td 48.6 & \sg 89.0 & \sg 9.0  & \sg 75.5 & \Tb 63.2 \\
 & VF -- XML         & \sg 89.8 & \sm 11.2 & \sm 65.3 & \Tc 53.6 & \sg 86.1 & \sg 9.3  & \sm 54.1 & \Td 42.6 & \sm 82.5 & \sm 12.0 & \sm 64.5 & \Td 48.3 & \sg 92.0 & \sg 6.0  & \sg 76.0 & \Tb 67.2 \\
 & CoT -- JSON       & \sm 82.1 & \sg 6.1  & \sg 66.8 & \Tc 52.3 & \sm 80.4 & \satb 15.5 & \sm 57.7 & \Td 40.2 & \satb 79.0 & \sm 13.0 & \sg 70.5 & \Tc 50.6 & \sg 92.5 & \sm 13.0 & \sg 81.5 & \Tb 69.4 \\
 & CoT -- XML        & \sm 84.2 & \sg 8.2  & \sg 65.8 & \Tc 52.0 & \sm 81.4 & \sm 12.4 & \sm 56.7 & \Td 41.1 & \satb 75.5 & \sm 12.0 & \sg 68.0 & \Td 46.8 & \sg 90.5 & \sm 11.1 & \sg 82.4 & \Tb 69.5 \\
\midrule
\multirow{5}{*}{\textbf{GPT-4o}}
 & Direct            & \sg 98.0 & \sm 11.2 & \sm 64.8 & \Tc 58.0 & \sg 95.5 & \sg 9.1  & \sm 58.6 & \Tc 51.6 & \sg 94.5 & \satb 24.0 & \satb 51.0 & \Te 36.9 & \sg 98.5 & \sg 8.0  & \sg 71.5 & \Tb 66.5 \\
 & VF -- JSON        & \sg 96.9 & \sm 11.2 & \sm 63.3 & \Tc 55.9 & \sg 96.0 & \sm 12.1 & \sm 61.1 & \Tc 52.8 & \sg 94.0 & \satb 31.0 & \satb 54.5 & \Te 36.7 & \sg 97.0 & \sg 7.0  & \sg 74.5 & \Tb 68.9 \\
 & VF -- XML         & \sg 98.0 & \sm 14.3 & \sm 62.8 & \Tc 54.5 & \sg 92.4 & \sm 14.1 & \sm 60.1 & \Td 49.0 & \sg 93.5 & \satb 35.0 & \satb 52.5 & \Te 32.7 & \sg 97.0 & \sg 6.0  & \sg 76.0 & \Ta 70.8 \\
 & CoT -- JSON       & \sg 89.8 & \satb 19.4 & \sm 61.2 & \Td 46.2 & \sg 92.4 & \satb 22.2 & \sm 60.1 & \Td 45.3 & \sm 88.0 & \satb 47.0 & \satb 54.5 & \Te 27.3 & \sg 92.5 & \satb 25.0 & \sg 70.5 & \Tc 53.7 \\
 & CoT -- XML        & \sg 90.8 & \satb 16.3 & \sm 61.2 & \Td 48.2 & \sg 92.9 & \satb 21.4 & \sm 59.2 & \Td 45.1 & \sm 88.0 & \satb 46.0 & \satb 53.5 & \Te 26.8 & \sg 93.0 & \satb 19.0 & \sg 73.5 & \Tb 59.5 \\
\midrule
\multirow{5}{*}{\textbf{Gemini-2.5-Pro}}
 & Direct            & \sm 87.8 & \satb 15.3 & \sm 62.8 & \Td 48.4 & \sm 84.0 & \satb 20.6 & \sm 55.2 & \Te 37.7 & \sm 81.5 & \satb 56.0 & \sm 63.5 & \Te 28.9 & \sm 87.0 & \satb 20.0 & \sg 80.0 & \Tb 60.9 \\
 & VF -- JSON        & \sm 87.0 & \satb 16.7 & \sm 62.2 & \Td 46.8 & \sm 87.1 & \sm 12.4 & \sm 60.3 & \Td 47.1 & \sg 91.1 & \sg 6.9  & \sg 88.5 & \Ta 77.5 & \sg 92.5 & \sm 13.0 & \sg 85.0 & \Ta 72.6 \\
 & VF -- XML         & \sm 88.8 & \satb 29.6 & \sm 59.7 & \Te 39.9 & \sm 84.5 & \satb 40.2 & \satb 54.6 & \Te 29.2 & \sm 83.8 & \satb 37.4 & \sg 80.3 & \Tc 51.6 & \sm 83.0 & \satb 33.0 & \sg 78.5 & \Tc 51.5 \\
 & CoT -- JSON       & \sm 87.6 & \sg 7.3  & \sm 62.7 & \Tc 51.7 & \sm 84.5 & \sm 14.4 & \sm 58.2 & \Td 43.1 & \sm 89.3 & \sg 5.4  & \sg 88.7 & \Ta 76.8 & \sm 89.5 & \sg 9.0  & \sg 85.5 & \Ta 72.5 \\
 & CoT -- XML        & \sm 88.5 & \satb 15.8 & \sm 60.4 & \Td 46.5 & \sm 82.5 & \satb 17.5 & \sm 56.2 & \Te 39.1 & \sm 89.9 & \sg 8.8  & \sg 89.9 & \Ta 76.9 & \sm 89.0 & \satb 17.0 & \sg 83.5 & \Tb 66.8 \\
\midrule
\multirow{5}{*}{\textbf{Gemini-2.0-Flash}}
 & Direct            & \sg 99.5 & \satb 16.3 & \sm 60.7 & \Tc 52.3 & \sg 99.5 & \satb 22.2 & \satb 52.3 & \Td 41.0 & \sg 99.0 & \satb 46.0 & \sm 55.5 & \Te 32.2 & \sg 98.0 & \satb 23.0 & \sm 61.5 & \Td 49.0 \\
 & VF -- JSON        & \sg 100.0& \satb 40.0 & \sm 55.4 & \Te 35.4 & \sg 99.5 & \satb 53.5 & \sm 59.3 & \Te 32.4 & \sg 98.5 & \satb 82.0 & \satb 52.0 & \Te 10.8 & \sg 98.0 & \satb 47.9 & \sm 63.8 & \Te 39.1 \\
 & VF -- XML         & \sg 99.0 & \satb 36.5 & \sm 59.3 & \Td 40.6 & \sg 97.5 & \satb 30.0 & \sm 61.5 & \Td 45.3 & \sg 100.0& \satb 98.0 & \satb 51.0 & \Te \phantom{0}2.0 & \sg 98.5 & \satb 42.4 & \sm 64.3 & \Td 42.5 \\
 & CoT -- JSON       & \sg 96.9 & \satb 17.0 & \sm 59.7 & \Td 49.6 & \sg 97.5 & \satb 19.0 & \sm 61.5 & \Tc 50.7 & \sg 96.0 & \satb 38.4 & \sm 56.8 & \Te 36.1 & \sg 96.9 & \satb 38.9 & \sm 63.4 & \Td 42.6 \\
 & CoT -- XML        & \sg 97.9 & \satb 26.8 & \sm 57.4 & \Td 43.1 & \sg 98.5 & \satb 31.3 & \sm 61.8 & \Td 45.5 & \sg 96.0 & \satb 50.0 & \sm 58.0 & \Te 31.7 & \sg 97.5 & \satb 39.4 & \sm 62.3 & \Td 41.5 \\
\midrule
\multirow{5}{*}{\textbf{Claude-Sonnet-4.6}}
 & Direct            & \sg 99.5 & \sg 7.7  & \sg 67.6 & \Tb 63.4 & \sg 97.9 & \sg 7.4  & \sm 54.7 & \Td 49.9 & \sg 98.8 & \sm 14.9 & \sm 57.3 & \Td 49.3 & \sg 99.5 & \sm 12.2 & \sg 79.7 & \Ta 73.2 \\
 & VF -- JSON        & \sg 98.4 & \sg 4.4  & \sg 68.5 & \Tb 65.2 & \sg 98.5 & \sg 7.2  & \sm 57.4 & \Tc 53.0 & \sg 98.1 & \sm 12.7 & \sg 71.8 & \Tb 64.2 & \sg 99.5 & \sm 14.1 & \sg 80.8 & \Ta 73.4 \\
 & VF -- XML         & \sg 91.8 & \sg 5.1  & \sg 67.3 & \Tb 59.4 & \sg 96.4 & \sg 8.2  & \sm 57.1 & \Tc 51.1 & \sm 89.9 & \sg 7.1  & \sg 83.4 & \Ta 71.8 & \sg 99.0 & \sm 12.0 & \sg 82.5 & \Ta 75.7 \\
 & CoT -- JSON       & \sm 89.4 & \sg 8.7  & \sg 70.2 & \Tb 58.9 & \sm 88.2 & \sg 7.2  & \sm 59.0 & \Td 48.9 & \sm 83.4 & \sg 7.1  & \sg 82.1 & \Tb 65.5 & \sg 95.5 & \sg 5.1  & \sg 87.4 & \Ta 81.0 \\
 & CoT -- XML        & \sm 84.2 & \sg 2.0  & \sg 67.9 & \Tc 56.3 & \sm 88.8 & \sm 10.2 & \sm 60.2 & \Td 48.9 & \sm 81.5 & \sg 3.0  & \sg 87.0 & \Tb 69.7 & \sg 92.0 & \sg 6.0  & \sg 87.0 & \Ta 77.3 \\
\midrule
\multirow{5}{*}{\textbf{Llama-3.3-70B}}
 & Direct            & \sg 100.0& \sm 13.3 & \sm 60.2 & \Tc 53.6 & \sg 100.0& \satb 16.3 & \sm 60.7 & \Tc 52.6 & \sg 100.0& \satb 26.0 & \sm 55.5 & \Td 42.5 & \sg 100.0& \sm 12.0 & \sm 63.5 & \Tc 57.5 \\
 & VF -- JSON        & \sg 100.0& \satb 15.3 & \sm 62.2 & \Tc 54.6 & \sg 100.0& \satb 17.3 & \sm 61.2 & \Tc 52.6 & \sg 100.0& \satb 22.4 & \satb 48.2 & \Te 37.0 & \sg 100.0& \satb 17.0 & \sm 64.0 & \Tc 55.5 \\
 & VF -- XML         & \sg 100.0& \sm 11.2 & \sm 62.2 & \Tc 56.6 & \sg 100.0& \sm 13.1 & \sm 59.6 & \Tc 53.1 & \sg 99.5 & \satb 37.0 & \satb 44.0 & \Te 25.4 & \sg 99.5 & \satb 15.0 & \sm 64.0 & \Tc 56.2 \\
 & CoT -- JSON       & \sg 98.0 & \sm 12.2 & \sm 62.8 & \Tc 55.6 & \sg 100.0& \satb 17.3 & \sm 62.4 & \Tc 53.8 & \sg 99.5 & \satb 45.5 & \satb 51.5 & \Te 28.6 & \sg 99.0 & \sm 14.0 & \sg 66.0 & \Tc 58.4 \\
 & CoT -- XML        & \sg 99.5 & \satb 16.3 & \sm 61.2 & \Tc 52.8 & \sg 98.0 & \satb 22.2 & \sm 62.1 & \Tc 50.0 & \sg 99.5 & \satb 44.0 & \satb 51.5 & \Te 29.4 & \sg 100.0& \satb 15.0 & \sg 66.5 & \Tb 59.0 \\
\bottomrule
\end{tabular}}
\caption{Stochastic consistency $C$ (\% deterministic across 10 reps, $\uparrow$), order flip rate $\varphi$ (\% verdict reversal on swap, $\downarrow$), accuracy $A$ (\%, $\uparrow$), and trustworthy verdict rate \color{blue}$T = C \cdot (A - \varphi/2)$ \normalcolor (in \%, $\uparrow$) at $T_s{=}0$. \textit{Direct}: single-token verdict. \textit{VF}: verdict before rationale. \textit{CoT}: rationale before verdict. $T$ shaded \colorbox{green!40}{$\geq$70} / \colorbox{green!18}{60--70} / \colorbox{yellow!25}{50--60} / \colorbox{orange!25}{40--50} / \colorbox{red!25}{$<$40}; support columns shaded in muted greys.}

\label{tab:consistency_full}
\end{table*}

\subsection{The Position-Bias Ceiling is Tight}
\label{sec:findings-ceiling}

Equation~\eqref{eq:ceiling} predicts that no judge can exceed observable accuracy $1-\varphi/2$ on a dataset with position flip rate $\varphi$. This bound is not merely formal: several cells in Table~\ref{tab:consistency_full} lie on, or within a few points of, the ceiling. The clearest example is Gemini-2.0-Flash with Verdict-first XML on JudgeBench. Its flip rate is $\varphi=98.0\%$, implying a maximum attainable accuracy of $51.0\%$, and the observed accuracy is exactly $A=51.0\%$. Under Equation~\eqref{eq:trustworthy}, this yields $T=2.0\%$: the judge is perfectly reproducible across repeated calls ($C=100\%$), but its verdicts carry essentially no trustworthy signal beyond order-induced coin-flipping.

\begin{figure}[h!]
  \centering
  \includegraphics[width=0.5\textwidth]{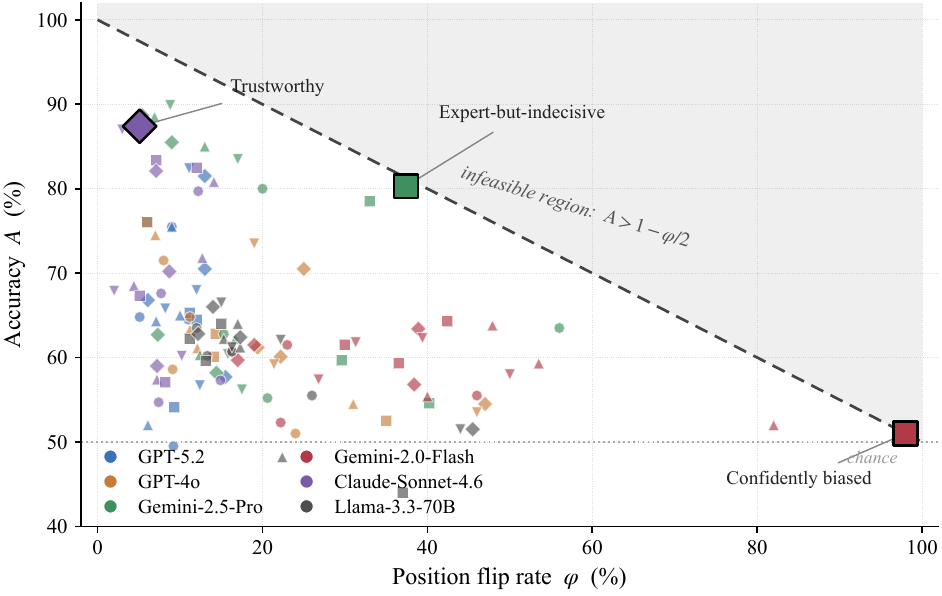}
  \caption{\textbf{Accuracy ceiling} view of the Trustworthy Verdict Rate at $T_s{=}0$.
  Each point is one cell of Table~\ref{tab:consistency_full}; the dashed line marks the upper bound
  $A \leq 1-\varphi/2$ from Equation~\eqref{eq:ceiling}. The shaded region is mathematically infeasible:
  position bias imposes an accuracy tax of $\varphi/2$ that no capability improvement can overcome.
  Points near the ceiling are ceiling-bound; points far below the ceiling, such as several Llama-3.3 cells,
  still have capability gaps where a stronger judge could help.}
  \label{fig:ceiling}
\end{figure}

The same pattern appears, though less extremely, for Gemini-2.0-Flash with Verdict-first JSON on JudgeBench: $\varphi=82.0\%$ implies a ceiling of $59.0\%$, while the observed accuracy is $52.0\%$ and $T$ falls to $10.8\%$. Figure~\ref{fig:ceiling} visualizes this ceiling across all model--format--dataset cells. Points near the boundary are not primarily capability-limited; their gains require reducing $\varphi$, not improving the base model.

\subsection{$T$ Reveals Reliability Differences Hidden by Accuracy}
\label{sec:findings-t}

Accuracy is the standard scalar for judge quality, but Equation~\eqref{eq:trustworthy} predicts that two judges with similar $A$ can have very different $T$ whenever they differ in stochastic consistency or position sensitivity. Table~\ref{tab:consistency_full} confirms this in two ways.

First, changing only the output format can alter trustworthiness far more than accuracy suggests. On JudgeBench, Gemini-2.5-Pro improves from $A=63.5\%, T=28.9\%$ under Direct prompting to $A=88.5\%, T=77.5\%$ under Verdict-first JSON. The 25\% accuracy gain is accompanied by a 49\% gain in $T$, because the format change both improves accuracy and reduces position flips from $\varphi=56.0\%$ to $\varphi=6.9\%$.

\begin{figure}[h!]
  \centering
  \includegraphics[width=\linewidth]{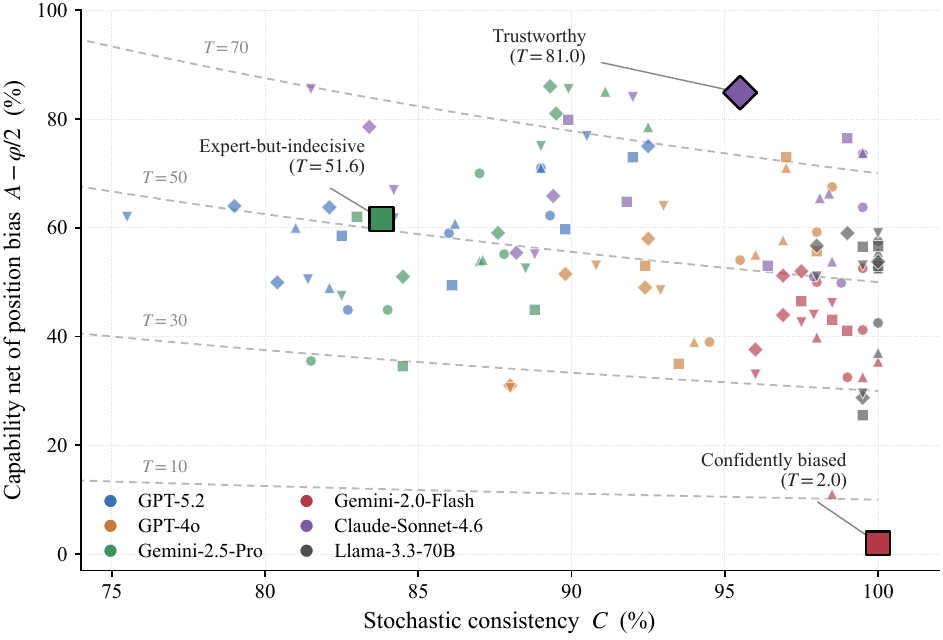}
  \caption{\textbf{Iso-$T$ factor space.}
  Cells from Table~\ref{tab:consistency_full} plotted in $(C,\; A-\varphi/2)$ coordinates, the two factors of
  $T=C \cdot (A-\varphi/2)$. Dashed curves are iso-$T$ contours. Highlighted examples show three distinct
  reliability regimes: \emph{Trustworthy} (Claude-Sonnet CoT-JSON / LLMBar, $T=81\%$),
  \emph{Expert-but-indecisive} (Gemini-2.5-Pro VF-XML / JudgeBench, $T=52\%$), and
  \emph{Confidently biased} (Gemini-2.0-Flash VF-XML / JudgeBench, $T=2\%$).}
  \label{fig:isoT}
\end{figure}

Second, comparable accuracies can mask different reliability profiles. On AlpacaFarm, GPT-5.2 with CoT-JSON and Claude-Sonnet-4.6 with Direct prompting achieve similar accuracies ($57.7\%$ vs.\ $54.7\%$), but their trustworthy rates differ substantially ($40.2\%$ vs.\ $49.9\%$). Claude's lower flip rate ($7.4\%$ vs.\ $15.5\%$) means that a larger share of its correct verdicts are reproducibly correct rather than order-contingent.

Figure~\ref{fig:isoT} shows the factorization directly by plotting cells in $(C, A-\varphi/2)$ space. The iso-$T$ contours make clear that trustworthiness can fail through either factor: low stochastic consistency, low bias-adjusted accuracy, or both.

\subsection{Consistency Alone Does Not Imply Trustworthiness}
\label{sec:findings-c-misleading}

A natural shortcut for reliability is to report stochastic consistency $C$: a judge that returns the same verdict on repeated calls feels dependable. Table~\ref{tab:consistency_full} shows why this intuition is incomplete. Llama-3.3-70B has $C \geq 98\%$ in every cell, yet its $T$ on JudgeBench ranges only from $25.4\%$ to $42.5\%$ because its flip rate remains high, ranging from $22\%$ to $45\%$. Gemini-2.0-Flash is even weaker: its $C$ is $96$--$100\%$ across settings, while $T$ spans $2.0\%$ to $52.3\%$.

The reason is structural. In Equation~\eqref{eq:trustworthy}, $C$ multiplies the bias-adjusted term $(A-\varphi/2)$. When this term is small, even perfect stochastic consistency cannot rescue trustworthiness. A deterministic judge can still be deterministically order-dependent.

\subsection{What Drives Variation in $T$?}
\label{sec:findings-variation}

The preceding results show that $T$ varies substantially across cells. We next decompose where that variation comes from: judge--dataset interaction, prompt format, and sampling temperature.

\paragraph{Judge--dataset interaction.}
A three-factor ANOVA of $T$ across the 120 judge--format--dataset cells shows that the largest source of variation is the judge\,$\times$\,dataset interaction ($\eta^2=28.1\%$), exceeding the judge ($27.0\%$) and dataset ($26.2\%$) main effects (see Appendix~\ref{app:tanova}). Thus, there is no universally best judge: rankings reorder across benchmarks. A complementary four-facet G-study over pair $\times$ judge $\times$ format $\times$ order further shows that order is a structured source of variance rather than residual noise, validating its explicit inclusion as $\varphi$ in $T$ (see Appendix~\ref{app:gstudy}).

\paragraph{Format dispersion.}
The dispersion $\Delta T_{\text{fmt}}$ from Equation~\eqref{eq:format-dispersion} measures how much trustworthiness depends on prompt format rather than judge identity. The largest dispersions concentrate on JudgeBench. Gemini-2.5-Pro spans $\Delta T_{\text{fmt}}=48.6$ points, from Direct prompting at $T=28.9\%$ to Verdict-first JSON at $T=77.5\%$; Gemini-2.0-Flash spans $34.1$ points. By contrast, Claude-Sonnet-4.6 on LLMBar varies only from $73.2\%$ to $81.0\%$ ($\Delta T_{\text{fmt}}=7.8$).

This means that, in high-dispersion settings, a single $(judge, format)$ result is not a stable estimate of the judge's underlying reliability. The format choice can carry as much signal as the model choice. Full details appear in Appendix~\ref{app:format-dispersion}, Table~\ref{tab:format_dispersion}.

\begin{figure}[h!]
  \centering
  \includegraphics[width=\linewidth]{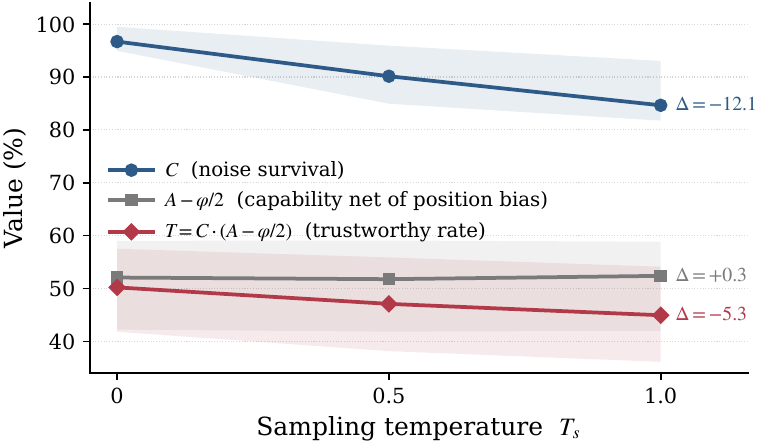}
  \caption{Mean $C$, $A-\varphi/2$, and $T$ across judge--format--dataset cells as sampling temperature
  $T_s$ varies from $0$ to $1.0$. Shaded bands show interquartile range. Temperature reduces $C$ by
  $12.1$ points but leaves $(A-\varphi/2)$ nearly unchanged, so the drop in $T$ is driven by noise survival,
  as Equation~\eqref{eq:trustworthy} predicts.}
  \label{fig:temperature_decomposition}
\end{figure}

\paragraph{Temperature ablation.}
Sampling temperature should primarily affect the noise-survival term $C$, not the bias-adjusted capability term $(A-\varphi/2)$. A sweep over $T_s\in\{0,0.5,1.0\}$ on the four temperature-tunable judges confirms this prediction. As shown in Figure~\ref{fig:temperature_decomposition}, mean $C$ drops by $12.1$ points across the sweep, while $(A-\varphi/2)$ changes by only $+0.3$ points across 48 cells. The resulting decline in $T$ is therefore attributable almost entirely to reduced stochastic consistency, supporting the interpretation that the two factors of $T$ capture distinct reliability properties. The complete temperature sweep analysis is presented in Appendix \ref{app:temperature}.

\subsection{Failure Modes and Practical Remedies}
\label{sec:findings-taxonomy}

\begin{figure}[h!]
  \centering
  \includegraphics[width=\linewidth]{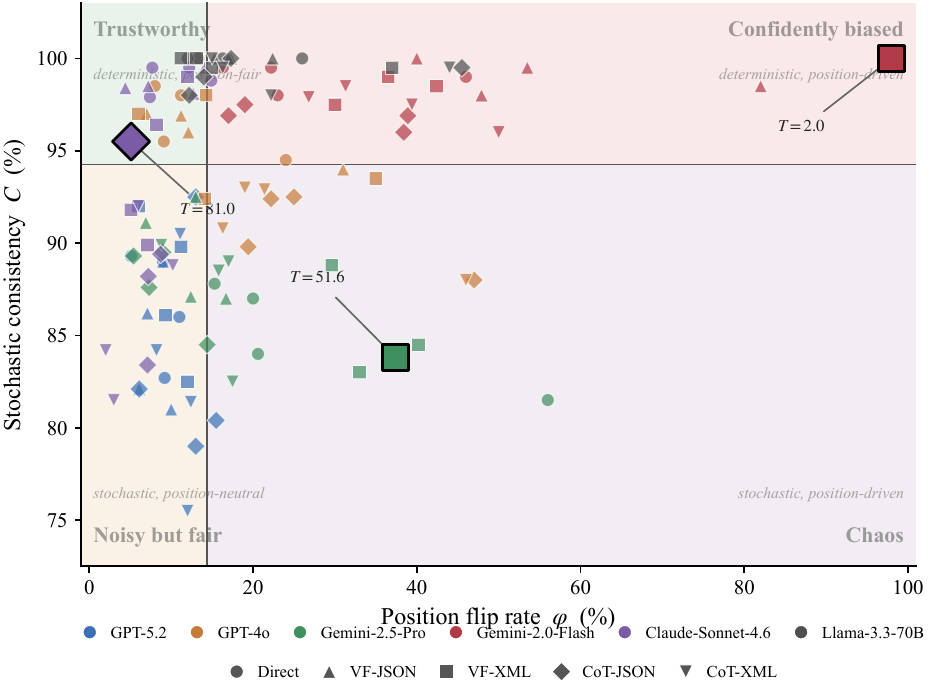}
  \caption{\textbf{Failure mode taxonomy.} Judge cells plotted by stochastic consistency $C$ and position flip rate $\varphi$.
  The quadrants distinguish trustworthy, noisy-but-fair, chaotic, and confidently biased judges, each requiring a different
  mitigation strategy.}
  \label{fig:taxonomy}
\end{figure}

Figure~\ref{fig:taxonomy} organizes judge failures into four regions of $(C,\varphi)$ space. The \emph{Trustworthy} region is the only regime where the residual gap $1-\varphi/2-A$ primarily reflects model capability rather than position bias; closing that gap requires a stronger judge. The \emph{Noisy but fair} region is easier to address: repeated sampling and majority vote can improve reliability because $\varphi$ is low. The \emph{Chaos} region requires both interventions: average over both candidate orderings to reduce position bias, then ensemble across repeated runs to reduce stochastic noise. The \emph{Confidently biased} region is the most dangerous. Here, high $C$ creates the appearance of reliability while high $\varphi$ drives $T$ toward zero. The worst observed cell in our study, with $\varphi=98\%$ and $T=2.0\%$, falls in this region.

\begin{figure}[h!]
  \centering
  \includegraphics[width=\columnwidth]{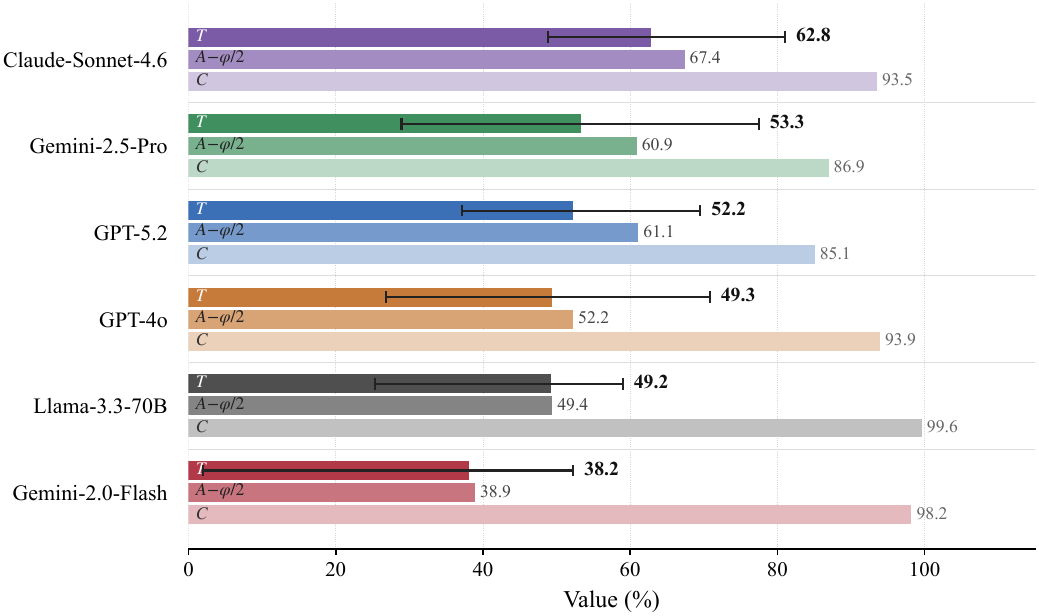}
  \caption{Per-judge mean of $C$, $A-\varphi/2$, and $T$ averaged across all 20 format--dataset cells per judge,
  ordered by mean $T$. Whiskers span the min--max range across cells. Wide ranges reflect judge\,$\times$\,(format, dataset)
  interactions but no judge dominates uniformly; see Appendix~\ref{app:gtheory}.}
  \label{fig:leaderboard}
\end{figure}

Figure~\ref{fig:leaderboard} aggregates these effects into a per-judge view. Sonnet-4.6 has the highest mean $T$, but the wide whiskers and overlap across judges show that no model dominates uniformly; reliability depends strongly on the dataset and format pairing.

\subsection{Rubric-Based Evaluation: FLASK and G-Eval}
\label{sec:findings-rubric}

\label{sec:results:rubric}

In the preceding sections, we characterize failures of direct pairwise evaluation. Here, we investigate if rubric-based paradigms can mitigate these failures. Specifically, can decomposing judgment into explicit criteria reduce $\varphi$ and thereby raise the $A_{\max}$ ceiling?

Table~\ref{tab:rubric} compares three paradigms on the two objective benchmarks, JudgeBench and LLMBar: Direct pairwise evaluation with one API call, FLASK-style \cite{ye2023flask} rubric decomposition with four calls per judgment, and G-Eval-style \cite{liu-etal-2023-g} score-based evaluation with one call over four scored criteria.

\begin{table*}[h!]
\centering
\resizebox{\textwidth}{!}{%
\begin{tabular}{l cccc cccc cccc cccc cccc cccc}
\toprule
\textbf{Judge} &
  \multicolumn{4}{c}{\textbf{JudgeBench -- Direct}} &
  \multicolumn{4}{c}{\textbf{JudgeBench -- FLASK}} &
  \multicolumn{4}{c}{\textbf{JudgeBench -- G-Eval}} &
  \multicolumn{4}{c}{\textbf{LLMBar -- Direct}} &
  \multicolumn{4}{c}{\textbf{LLMBar -- FLASK}} &
  \multicolumn{4}{c}{\textbf{LLMBar -- G-Eval}} \\
\cmidrule(lr){2-5}\cmidrule(lr){6-9}\cmidrule(lr){10-13}
\cmidrule(lr){14-17}\cmidrule(lr){18-21}\cmidrule(lr){22-25}
 & $C$ & $\varphi$ & $A$ & $T$
 & $C$ & $\varphi$ & $A$ & $T$
 & $C$ & $\varphi$ & $A$ & $T$
 & $C$ & $\varphi$ & $A$ & $T$
 & $C$ & $\varphi$ & $A$ & $T$
 & $C$ & $\varphi$ & $A$ & $T$ \\
\midrule
GPT-5.2 &
  \sm 86.0 & \sm 11.0 & \sm 64.5 & \Tc 50.7 &
  \satb 34.8 & \sg  9.1 & \sm 63.6 & \Te 20.5 &
  \satb 68.5 & \satb 16.0 & \sg 68.0 & \Td 41.1 &
  \sm 89.0 & \sg  9.0 & \sg 75.5 & \Tb 63.2 &
  \satb 55.0 & \satb 21.0 & \sg 79.5 & \Te 37.9 &
  \sm 88.0 & \satb 18.0 & \sg 79.0 & \Tb 61.6 \\
GPT-4o &
  \sg 94.5 & \satb 24.0 & \satb 51.0 & \Te 36.9 &
  \satb 44.4 & \satb 47.5 & \satb 52.5 & \Te 12.8 &
  \sm 89.0 & \satb 61.0 & \sm 62.0 & \Te 28.0 &
  \sg 98.5 & \sg  8.0 & \sg 71.5 & \Tb 66.5 &
  \satb 53.0 & \satb 31.0 & \sg 75.0 & \Te 31.5 &
  \sg 95.5 & \satb 28.0 & \sg 78.0 & \Tb 61.1 \\
Gemini-2.5-Pro &
  \sm 81.5 & \satb 56.0 & \sm 63.5 & \Te 28.9 &
  \satb 60.6 & \sm 10.5 & \sg 67.4 & \Te 37.7 &
  \sm 87.7 & \sg 10.0 & \sg 93.8 & \Ta \textbf{77.9} &
  \sm 87.0 & \satb 20.0 & \sg 80.0 & \Tb 60.9 &
  \satb 63.0 & \sm 14.0 & \sg 78.0 & \Td 44.7 &
  \sm 86.0 & \satb 27.0 & \sg 79.0 & \Tc 56.3 \\
Gemini-2.0-Flash &
  \sg 99.0 & \satb 46.0 & \sm 55.5 & \Te 32.2 &
  \sg 90.4 & \satb 30.3 & \sm 57.1 & \Te 37.9 &
  \sg 94.5 & \satb 40.4 & \sm 59.0 & \Te 36.7 &
  \sg 98.0 & \satb 23.0 & \sm 61.5 & \Td 49.0 &
  \satb 66.0 & \satb 32.0 & \sg 65.5 & \Te 32.7 &
  \sg 99.0 & \satb 39.0 & \sg 74.0 & \Tc 54.0 \\
Claude-Sonnet-4.6 &
  \sg 98.8 & \sm 14.9 & \sm 57.3 & \Td 49.3 &
  \satb 69.0 & \sg  2.0 & \sg 75.1 & \Tc 51.1 &
  \sm 78.7 & \sm 14.4 & \sg 82.0 & \Tc 58.8 &
  \sg 99.5 & \sm 12.2 & \sg 79.7 & \Ta \textbf{73.2} &
  \satb 58.0 & \sm 11.0 & \sg 81.0 & \Td 43.8 &
  \sg 91.0 & \sg  9.1 & \sg 85.0 & \Ta \textbf{73.2} \\
Llama-3.3-70B &
  \sg 100.0 & \satb 26.0 & \sm 55.5 & \Td 42.5 &
  \sg  96.9 & \satb 51.0 & \satb 53.6 & \Te 27.2 &
  \sg  97.0 & \satb 55.1 & \satb 52.0 & \Te 23.7 &
  \sg 100.0 & \sm 12.0 & \sm 63.5 & \Tc 57.5 &
  \sg  97.5 & \satb 20.0 & \sg 67.5 & \Tc 56.1 &
  \sg 100.0 & \satb 29.0 & \sg 78.0 & \Tb 63.5 \\
\bottomrule
\end{tabular}}
\caption{Three-paradigm comparison at $T_s{=}0$: Direct pairwise (1 API call), FLASK rubric decomposition (4 calls per judgment), and G-Eval score-based evaluation (1 call with 4 scored criteria). All values in percentage points. $T = C \cdot (A - \varphi/2)$ shaded \colorbox{green!40}{$\geq$70} / \colorbox{green!18}{60--70} / \colorbox{yellow!25}{50--60} / \colorbox{orange!25}{40--50} / \colorbox{red!25}{$<$40}; support columns ($C$, $\varphi$, $A$) in muted greys. \textbf{Bold} = best $T$ per dataset.}
\label{tab:rubric}
\end{table*}

FLASK reduces $\varphi$ for some judges but collapses $C$ universally (e.g., GPT-5.2: $C\,86.0\% \to 34.8\%$ on JudgeBench), leaving $T$ lower than Direct for five of six judges.
G-Eval elicits all criterion scores in a single prompt, preserving $C$ while reducing position sensitivity; for capable judges this can lift $T$ substantially (Gemini-2.5-Pro: $T\,28.9\% \to 77.9\%$ on JudgeBench, a $+49$ pp gain exceeding any format-induced swing in Table~\ref{tab:format_dispersion}).

\section{Conclusion and Recommendations}
\label{sec:conclusion}

We have audited six frontier LLM judges across four benchmarks,
five prompt formats, two presentation orders, three sampling
temperatures, and ten repetitions, with an additional comparison
of direct, FLASK-decomposed, and G-Eval rubric paradigms on the
two objective benchmarks. The audit yields that judge reliability is not what accuracy
measures. A verdict is operationally \textit{trustworthy} only when it is
simultaneously \textbf{reproducible} across runs, \textbf{invariant to
candidate-order swap}, and \textbf{correct}. We formalize this as the
\emph{trustworthy verdict rate}, and prove that the same
formalization implies a hard capability ceiling that no improvement in model capability
can overcome at fixed position flip rate. Both quantities are
directly observable from repeated elicitations and require no
ground-truth labels beyond what existing benchmarks already
provide.

We also argue that this ceiling is empirically
tight; i.e.,
near-deterministic judges can still produce verdicts that are
worse than chance; and reliability is item-specific rather than
judge-level. The most striking case
in our data is a single judge $\times$ format $\times$  dataset configuration
with $C = 100\%$ and $T = 2\%$: a judge that appears maximally
reliable by every single-shot diagnostic in common use, but
contributes essentially nothing beyond order-induced coin-flipping.

The implication for practice is concrete. As LLM judges become
infrastructure for RLHF, leaderboards, automated evaluation
pipelines, and downstream model selection --- the field needs
reliability audits alongside accuracy benchmarks. We recommend
that benchmark releases report at minimum $(C, \varphi, A, T)$
per cell, that judge selection use $T$ rather than accuracy as
the comparative scalar, and that protocols which leave $\varphi$
unobservable (single-order evaluation) be treated as
fundamentally under-specified. None of these recommendations
require new infrastructure; they require reading existing
infrastructure with a richer lens than the one currently in use.

The trustworthy verdict rate is not a competitor to accuracy.
It is what accuracy was always meant to mean.

\section*{Limitations}
Our study has three primary limitations. First, due to computational tractability, we restricted our evaluation to four English-only benchmarks. Consequently, our reliability findings may not generalize to low-resource languages where frontier models exhibit weaker baseline performance, and our estimates for low-frequency phenomena, such as per-domain flip rates, carry minor sampling variance that headline numbers may not surface. Second, we evaluate closed-weight models without access to log-probabilities or internal states, which precludes the analysis of finer-grained token confidence signals. Third, our audit focuses exclusively on pairwise preference data. While pairwise evaluation is currently the most prevalent paradigm, a natural extension of this work would be to investigate whether our findings regarding verdict instability and bias transfer to list-wise ranking or multi-model grading frameworks.
\bibliography{custom}

\appendix
\onecolumn

  \section*{Appendix Contents}
\label{app:toc}

{%
  \vspace{-0.4em}\hrule height 0.4pt\vspace{0.6em}
  \small
  \newcommand{\tocsec}[2]{%
    \noindent\textbf{\hyperref[#1]{#2}}%
    \nobreak\hfill\textbf{\pageref{#1}}\par\vspace{0.15em}}
  \newcommand{\tocsub}[2]{%
    \noindent\hspace{1.4em}\hyperref[#1]{#2}%
    \nobreak\hfill\pageref{#1}\par\vspace{0.05em}}
  \tocsec{app:temperature}{A\quad Temperature Sweep Analysis}
  \tocsec{app:format-dispersion}{B\quad Format Dispersion Analysis}
  \tocsec{app:gtheory}{C\quad Variance Decomposition Details}
    \tocsub{app:tanova}{C.1\quad ANOVA on $T$}
    \tocsub{app:gstudy}{C.2\quad Four-Facet G-Study on Verdict Direction}

\tocsec{app:prompts}{D\quad Prompt Templates}
\tocsec{app:cost}{E\quad API Cost}
\tocsec{app:datasets}{F\quad Dataset Details}
\tocsec{app:a2_clarification}{G\quad Assumptions underlying $T$}
  \vspace{0.3em}\hrule height 0.4pt
}
\vspace{0.8em}
\section{Temperature Sweep Analysis}
\label{app:temperature}
Table~\ref{tab:temperature_sweep} reports $(C, \varphi, A, T)$ across
sampling temperatures $T_s \in \{0, 0.5, 1.0\}$ for the four
temperature-tunable judges (GPT-4o, Gemini-2.0-Flash,
Claude-Sonnet-4.6, Llama-3.3-70B; GPT-5.2 and Gemini-2.5-Pro are
excluded because their public APIs do not expose a temperature
parameter). Verdict-first and CoT rows aggregate over their JSON and
XML variants for compactness, collapsing the 5-format grid of
Table~\ref{tab:consistency_full} to a 3-format view.

\begin{table*}[h!]
\centering
\resizebox{\textwidth}{!}{%
\begin{tabular}{cl l cccc cccc cccc cccc}
\toprule
$T_s$ & \textbf{Judge} & \textbf{Format} & \multicolumn{4}{c}{\textbf{MT-Bench}} & \multicolumn{4}{c}{\textbf{AlpacaFarm}} & \multicolumn{4}{c}{\textbf{JudgeBench}} & \multicolumn{4}{c}{\textbf{LLMBar}} \\
\cmidrule(lr){4-7}\cmidrule(lr){8-11}\cmidrule(lr){12-15}\cmidrule(lr){16-19}
 & & & C & $\varphi$ & A & $T$ & C & $\varphi$ & A & $T$ & C & $\varphi$ & A & $T$ & C & $\varphi$ & A & $T$ \\
\midrule
\multirow{12}{*}{\rotatebox[origin=c]{90}{\textbf{$T=0$}}}
 & \multirow{3}{*}{GPT-4o}
   & Direct        & \sg 98.0 & \sm 11.2 & \sm 64.8 & \Tc 58.0 & \sg 95.5 & \sg  9.1 & \sm 58.6 & \Tc 51.6 & \sg 94.5 & \satb 24.0 & \satb 51.0 & \Te 36.9 & \sg 98.5 & \sg  8.0 & \sg 71.5 & \Tb 66.5 \\
 & & Verdict-first & \sg 97.4 & \sm 12.8 & \sm 63.0 & \Tc 55.1 & \sg 94.2 & \sm 13.1 & \sm 60.6 & \Tc 50.9 & \sg 93.8 & \satb 33.0 & \sm 53.5 & \Te 34.7 & \sg 97.0 & \sg  6.5 & \sg 75.2 & \Tb 69.8 \\
 & & CoT           & \sg 90.3 & \satb 17.9 & \sm 61.2 & \Td 47.2 & \sg 92.6 & \satb 21.8 & \sm 59.6 & \Td 45.1 & \sm 88.0 & \satb 46.5 & \sm 54.0 & \Te 27.1 & \sg 92.8 & \satb 22.0 & \sg 72.0 & \Tc 56.6 \\
\cmidrule(l){2-19}
 & \multirow{3}{*}{Gemini-2.0-Flash}
   & Direct        & \sg 99.5 & \satb 16.3 & \sm 60.7 & \Tc 52.3 & \sg 99.5 & \satb 22.2 & \sm 52.3 & \Td 41.0 & \sg 99.0 & \satb 46.0 & \sm 55.5 & \Te 32.2 & \sg 98.0 & \satb 23.0 & \sm 61.5 & \Td 49.0 \\
 & & Verdict-first & \sg 99.5 & \satb 38.2 & \sm 57.4 & \Te 38.1 & \sg 98.5 & \satb 41.8 & \sm 60.4 & \Te 38.9 & \sg 99.2 & \satb 90.0 & \sm 51.5 & \Te \phantom{0}6.4 & \sg 98.2 & \satb 45.2 & \sm 64.0 & \Td 40.7 \\
 & & CoT           & \sg 97.4 & \satb 21.9 & \sm 58.6 & \Td 46.4 & \sg 98.0 & \satb 25.2 & \sm 61.7 & \Td 48.1 & \sg 96.0 & \satb 44.2 & \sm 57.4 & \Te 33.9 & \sg 97.2 & \satb 39.2 & \sm 62.9 & \Td 42.1 \\
\cmidrule(l){2-19}
 & \multirow{3}{*}{Claude-Sonnet-4.6}
   & Direct        & \sg 99.5 & \sg  7.7 & \sg 67.6 & \Tb 63.4 & \sg 97.9 & \sg  7.4 & \sm 54.7 & \Td 49.9 & \sg 98.8 & \sm 14.9 & \sm 57.3 & \Td 49.3 & \sg 99.5 & \sm 12.2 & \sg 79.7 & \Ta 73.2 \\
 & & Verdict-first & \sg 95.1 & \sg  4.8 & \sg 67.9 & \Tb 62.3 & \sg 97.4 & \sg  7.7 & \sm 57.3 & \Tc 52.1 & \sg 94.0 & \sg  9.9 & \sg 77.6 & \Tb 68.3 & \sg 99.2 & \sm 13.1 & \sg 81.7 & \Ta 74.5 \\
 & & CoT           & \sm 86.8 & \sg  5.4 & \sg 69.0 & \Tc 57.5 & \sm 88.5 & \sg  8.7 & \sm 59.6 & \Td 48.9 & \sm 82.5 & \sg  5.1 & \sg 84.6 & \Tb 67.7 & \sg 93.7 & \sg  5.6 & \sg 87.2 & \Ta 79.1 \\
\cmidrule(l){2-19}
 & \multirow{3}{*}{Llama-3.3-70B}
   & Direct        & \sg 100  & \sm 13.3 & \sm 60.2 & \Tc 53.6 & \sg 100  & \satb 16.3 & \sm 60.7 & \Tc 52.6 & \sg 100  & \satb 26.0 & \sm 55.5 & \Td 42.5 & \sg 100  & \sm 12.0 & \sm 63.5 & \Tc 57.5 \\
 & & Verdict-first & \sg 100  & \sm 13.3 & \sm 62.2 & \Tc 55.6 & \sg 100  & \satb 15.2 & \sm 60.4 & \Tc 52.8 & \sg 99.8 & \satb 29.7 & \satb 46.1 & \Te 31.2 & \sg 99.8 & \satb 16.0 & \sm 64.0 & \Tc 55.9 \\
 & & CoT           & \sg 98.7 & \sm 14.3 & \sm 62.0 & \Tc 54.1 & \sg 99.0 & \satb 19.8 & \sm 62.3 & \Tc 51.9 & \sg 99.5 & \satb 44.7 & \sm 51.5 & \Te 29.0 & \sg 99.5 & \sm 14.5 & \sg 66.2 & \Tc 58.7 \\
\midrule
\multirow{12}{*}{\rotatebox[origin=c]{90}{\textbf{$T=0.5$}}}
 & \multirow{3}{*}{GPT-4o}
   & Direct        & \sg 91.9 & \sg  8.6 & \sg 65.1 & \Tc 55.9 & \sg 91.9 & \sm 11.8 & \sm 56.5 & \Td 46.5 & \satb 77.5 & \satb 24.0 & \satb 49.5 & \Te 29.1 & \sg 97.9 & \sg  7.4 & \sg 71.1 & \Tb 66.0 \\
 & & Verdict-first & \sg 93.0 & \sm 13.5 & \sm 62.4 & \Tc 51.8 & \sm 89.0 & \sm 13.4 & \sm 59.1 & \Td 46.6 & \satb 75.2 & \satb 29.1 & \sm 52.9 & \Te 28.8 & \sg 95.2 & \sg  5.6 & \sg 75.5 & \Tb 69.2 \\
 & & CoT           & \sm 84.8 & \satb 20.1 & \sm 61.7 & \Td 43.8 & \sm 84.4 & \satb 21.5 & \sm 59.9 & \Td 41.5 & \satb 62.1 & \satb 44.7 & \sm 52.5 & \Te 18.7 & \sm 89.9 & \satb 18.4 & \sg 72.0 & \Tc 56.5 \\
\cmidrule(l){2-19}
 & \multirow{3}{*}{Gemini-2.0-Flash}
   & Direct        & \sg 96.3 & \sm 12.9 & \sm 60.4 & \Tc 52.0 & \sg 91.9 & \satb 24.7 & \satb 50.0 & \Te 34.6 & \sm 85.0 & \satb 46.0 & \sm 55.5 & \Te 27.6 & \sg 94.9 & \satb 25.0 & \sm 62.2 & \Td 47.2 \\
 & & Verdict-first & \sg 96.0 & \satb 37.3 & \sm 57.9 & \Te 37.7 & \sg 93.3 & \satb 43.0 & \sm 59.4 & \Te 35.4 & \sg 95.8 & \satb 90.0 & \sm 51.5 & \Te \phantom{0}6.2 & \sg 94.6 & \satb 47.2 & \sm 64.5 & \Te 38.7 \\
 & & CoT           & \sm 83.2 & \satb 19.1 & \sm 58.5 & \Td 40.7 & \sm 80.9 & \satb 24.7 & \sm 59.7 & \Te 38.3 & \satb 62.8 & \satb 43.0 & \sm 57.5 & \Te 22.6 & \sm 82.3 & \satb 39.5 & \sm 63.7 & \Te 36.2 \\
\cmidrule(l){2-19}
 & \multirow{3}{*}{Claude-Sonnet-4.6}
   & Direct        & \sg 98.3 & \sg  8.0 & \sg 67.6 & \Tb 62.5 & \sg 95.1 & \sg  5.5 & \sm 54.1 & \Td 48.8 & \sg 94.8 & \satb 16.2 & \sm 55.2 & \Td 44.7 & \sg 95.9 & \sm 12.6 & \sg 79.0 & \Tb 69.7 \\
 & & Verdict-first & \sg 94.8 & \sg  5.1 & \sg 68.3 & \Tb 62.3 & \sg 94.9 & \sg  8.1 & \sm 57.7 & \Tc 50.9 & \sg 91.9 & \sg  9.9 & \sg 77.6 & \Tb 66.8 & \sg 99.0 & \sm 13.7 & \sg 81.5 & \Ta 73.9 \\
 & & CoT           & \sm 84.3 & \sg  5.1 & \sg 68.8 & \Tc 55.8 & \sm 87.0 & \sm 10.9 & \sm 58.9 & \Td 46.5 & \satb 79.7 & \sg  6.3 & \sg 85.2 & \Tb 65.4 & \sg 93.9 & \sg  6.7 & \sg 87.1 & \Ta 78.6 \\
\cmidrule(l){2-19}
 & \multirow{3}{*}{Llama-3.3-70B}
   & Direct        & \sg 98.9 & \sg  9.8 & \sm 61.4 & \Tc 55.9 & \sg 98.4 & \satb 17.2 & \sm 58.6 & \Td 49.2 & \sg 97.5 & \satb 26.3 & \sm 55.3 & \Td 41.1 & \sg 98.5 & \sm 11.0 & \sm 64.0 & \Tc 57.6 \\
 & & Verdict-first & \sg 97.9 & \sm 11.8 & \sm 62.7 & \Tc 55.6 & \sg 97.3 & \satb 16.7 & \sm 58.6 & \Td 48.9 & \sm 89.2 & \satb 29.2 & \satb 46.5 & \Te 28.5 & \sg 97.2 & \satb 16.5 & \sm 63.8 & \Tc 54.0 \\
 & & CoT           & \sg 90.9 & \sm 13.9 & \sm 61.2 & \Td 49.3 & \sg 90.9 & \satb 19.9 & \sm 60.8 & \Td 46.2 & \satb 78.5 & \satb 46.0 & \sm 51.8 & \Te 22.6 & \sg 92.8 & \sm 15.0 & \sg 65.8 & \Tc 54.1 \\
\midrule
\multirow{12}{*}{\rotatebox[origin=c]{90}{\textbf{$T=1.0$}}}
 & \multirow{3}{*}{GPT-4o}
   & Direct        & \sm 87.1 & \sg  8.2 & \sg 66.5 & \Tc 54.4 & \sm 83.5 & \sg  8.0 & \sm 58.0 & \Td 45.1 & \satb 58.6 & \satb 22.2 & \satb 49.0 & \Te 22.2 & \sg 93.0 & \sm 10.0 & \sg 72.0 & \Tb 62.3 \\
 & & Verdict-first & \sm 88.1 & \sm 13.9 & \sm 63.7 & \Tc 50.0 & \sm 82.2 & \sm 10.5 & \sm 60.5 & \Td 45.4 & \satb 58.8 & \satb 31.3 & \sm 53.8 & \Te 22.4 & \sg 90.2 & \sg  6.0 & \sg 74.9 & \Tb 64.9 \\
 & & CoT           & \sm 82.0 & \satb 21.6 & \sm 62.9 & \Td 42.7 & \satb 75.8 & \satb 18.5 & \sm 60.2 & \Te 38.6 & \satb 50.0 & \satb 35.9 & \sm 54.0 & \Te 18.0 & \sm 81.0 & \satb 17.0 & \sg 72.5 & \Tc 51.8 \\
\cmidrule(l){2-19}
 & \multirow{3}{*}{Gemini-2.0-Flash}
   & Direct        & \sg 93.3 & \satb 15.5 & \sm 61.3 & \Tc 50.0 & \sm 88.0 & \satb 22.0 & \sm 53.5 & \Te 37.4 & \satb 78.0 & \satb 46.0 & \sm 55.5 & \Te 25.4 & \sm 89.0 & \satb 25.0 & \sm 62.0 & \Td 44.1 \\
 & & Verdict-first & \sg 94.3 & \satb 37.7 & \sm 57.9 & \Te 36.8 & \sm 88.5 & \satb 41.0 & \sm 60.8 & \Te 35.7 & \sg 93.7 & \satb 89.9 & \sm 52.0 & \Te \phantom{0}6.6 & \sm 87.7 & \satb 47.5 & \sm 63.1 & \Te 34.5 \\
 & & CoT           & \satb 75.9 & \satb 19.0 & \sm 57.3 & \Te 36.3 & \satb 63.8 & \satb 26.5 & \sm 60.8 & \Te 30.3 & \satb 50.8 & \satb 43.9 & \sm 59.8 & \Te 19.2 & \satb 62.5 & \satb 40.3 & \sg 67.8 & \Te 29.8 \\
\cmidrule(l){2-19}
 & \multirow{3}{*}{Claude-Sonnet-4.6}
   & Direct        & \sg 95.1 & \sg  8.7 & \sg 66.5 & \Tb 59.1 & \sg 90.5 & \sg  8.1 & \sm 54.8 & \Td 45.9 & \sm 89.4 & \satb 19.5 & \sm 56.4 & \Td 41.7 & \sg 92.5 & \sm 13.1 & \sg 79.4 & \Tb 67.4 \\
 & & Verdict-first & \sg 93.2 & \sg  4.9 & \sg 67.7 & \Tb 60.8 & \sg 92.5 & \sg  8.0 & \sm 57.2 & \Td 49.2 & \sm 89.5 & \sg  9.7 & \sg 77.8 & \Tb 65.3 & \sg 97.5 & \sm 12.6 & \sg 81.2 & \Ta 73.0 \\
 & & CoT           & \sm 83.6 & \sg  3.3 & \sg 68.5 & \Tc 55.9 & \sm 83.0 & \sg  8.0 & \sm 59.0 & \Td 45.6 & \satb 77.5 & \sg  4.6 & \sg 85.0 & \Tb 64.1 & \sg 90.2 & \sg  6.0 & \sg 87.5 & \Ta 76.2 \\
\cmidrule(l){2-19}
 & \multirow{3}{*}{Llama-3.3-70B}
   & Direct        & \sg 99.0 & \sm 12.4 & \sm 60.8 & \Tc 54.1 & \sg 96.0 & \satb 17.0 & \sm 61.0 & \Tc 50.4 & \sg 93.0 & \satb 28.0 & \sm 54.5 & \Te 37.7 & \sg 98.0 & \sm 12.0 & \sm 63.5 & \Tc 56.4 \\
 & & Verdict-first & \sg 94.6 & \sm 14.0 & \sm 62.8 & \Tc 52.8 & \sg 94.8 & \sm 15.0 & \sm 60.8 & \Tc 50.5 & \sm 83.7 & \satb 31.1 & \satb 46.4 & \Te 25.8 & \sg 93.2 & \satb 17.0 & \sm 64.0 & \Tc 51.7 \\
 & & CoT           & \sm 89.4 & \sm 13.5 & \sm 62.7 & \Tc 50.0 & \sm 89.2 & \satb 19.0 & \sm 62.0 & \Td 46.8 & \satb 71.8 & \satb 44.0 & \sm 51.0 & \Te 20.8 & \sg 90.2 & \sm 15.5 & \sg 65.5 & \Tc 52.1 \\
\bottomrule
\end{tabular}}
\caption{Stochastic consistency $C$, position flip rate $\varphi$, accuracy $A$, and trustworthy rate $T = C \cdot (A - \varphi/2)$ (all \%) across sampling temperatures $T_s \in \{0, 0.5, 1.0\}$. Verdict-first and CoT rows are averages over JSON and XML format variants. $T$ shaded \colorbox{green!40}{$\geq$70} / \colorbox{green!18}{60--70} / \colorbox{yellow!25}{50--60} / \colorbox{orange!25}{40--50} / \colorbox{red!25}{$<$40}; support columns shaded in muted greys.}
\label{tab:temperature_sweep}
\end{table*}
\newpage
\section{Format Dispersion Analysis}
\label{app:format-dispersion}

To quantify the sensitivity of a judge to prompt formatting, we define the
\emph{format dispersion} of the trustworthy rate as
\[
\Delta T_{\text{fmt}}(J, D)
=
\max_f T(J, f, D)
-
\min_f T(J, f, D),
\]
where $J$ denotes the judge model, $D$ the benchmark dataset, and $f$ the prompting format. Intuitively, $\Delta T_{\text{fmt}}$ measures how much the observed trustworthiness of a judge changes solely due to formatting choices while keeping the judge and evaluation data fixed. A low value indicates that the judge is robust to presentation format, whereas a high value indicates strong format dependence.

Table~\ref{tab:format_dispersion} reports $\Delta T_{\text{fmt}}(J, D)$ together with the corresponding worst-case trustworthy rate
\[
T_{\min}(J, D)
=
\min_f T(J, f, D).
\]

\begin{table}[h!]
\centering
\small

\newcommand{\Da}{\cellcolor{green!40}}   
\newcommand{\Db}{\cellcolor{green!18}}   
\newcommand{\Dc}{\cellcolor{yellow!25}}  
\newcommand{\Dd}{\cellcolor{orange!25}}  
\newcommand{\De}{\cellcolor{red!25}}     

\begin{tabular}{l cccc c}
\toprule
\textbf{Judge} & \textbf{MT Bench} & \textbf{AlpacaFarm} & \textbf{JudgeBench} & \textbf{LLMBar} & \textbf{Mean} \\
\midrule
GPT-5.2            & \Da $3.6_{\,52.0}$  & \Db $5.5_{\,37.1}$  & \Da $3.9_{\,46.8}$  & \Db $6.3_{\,63.2}$  & 4.8 \\
GPT-4o             & \Dc $11.8_{\,46.2}$ & \Db $7.7_{\,45.1}$  & \Dc $10.1_{\,26.8}$ & \Dc $17.1_{\,53.7}$ & 11.7 \\
Gemini-2.5-Pro     & \Dc $11.8_{\,39.9}$ & \Dc $17.9_{\,29.2}$ & \De $48.6_{\,28.9}$ & \Dd $21.1_{\,51.5}$ & 24.9 \\
Gemini-2.0-Flash   & \Dc $16.9_{\,35.4}$ & \Dc $18.3_{\,32.4}$ & \Dd $34.1_{\,\phantom{0}2.0}$ & \Db $9.9_{\,39.1}$ & 19.8 \\
Claude-Sonnet-4.6  & \Db $8.9_{\,56.3}$  & \Da $4.1_{\,48.9}$  & \Dd $22.5_{\,49.3}$ & \Db $7.8_{\,73.2}$  & 10.8 \\
Llama-3.3-70B      & \Da $3.8_{\,52.8}$  & \Da $3.8_{\,50.0}$  & \Dc $17.1_{\,25.4}$ & \Da $3.5_{\,55.5}$  & 7.1 \\
\midrule
\textbf{Mean}      & 9.5 & 9.6 & 22.7 & 10.8 & \\
\bottomrule
\end{tabular}

\caption{
Format-induced dispersion of the trustworthy rate,
$\Delta T_{\text{fmt}}(J, D)$, with the worst-case trustworthy rate
$T_{\min}$ shown in subscript.
Shading indicates dispersion magnitude:
\colorbox{green!40}{$<$5},
\colorbox{green!18}{5--10},
\colorbox{yellow!25}{10--20},
\colorbox{orange!25}{20--35},
and
\colorbox{red!25}{$>$35}.
Low $\Delta T_{\text{fmt}}$ indicates a format-robust judge. All values are reported in percentage points.
}
\label{tab:format_dispersion}
\end{table}

\subsection*{Findings} Two broad patterns emerge from Table~\ref{tab:format_dispersion}.

\paragraph{Format robustness varies substantially across judges.}
GPT-5.2 and Llama-3.3-70B exhibit consistently low dispersion across benchmarks, suggesting that their trustworthiness is largely invariant to prompt format. In contrast, the two Gemini variants show substantially larger swings, with mean dispersions of $24.9\%$ and $19.8\%$  respectively. The most extreme case occurs for Gemini-2.5-Pro on JudgeBench, where the trustworthy rate ranges from $28.9\%$ under Direct prompting to $77.5\%$ under VF-JSON, yielding a dispersion of $48.6$ points attributable solely to formatting changes.

\paragraph{JudgeBench is markedly more format-sensitive.}
Average dispersion across judges is approximately $10$ points on MT-Bench, AlpacaFarm, and LLMBar, but rises to $22.7\%$ on JudgeBench. Moreover, five of the six largest dispersion values occur on JudgeBench, and the only configuration with $T_{\min} < 10\%$ appears on this benchmark. One possible explanation is that harder or more knowledge-intensive comparisons weaken the judge's underlying confidence signal, thereby amplifying the effect of surface-level prompt presentation. We leave systematic investigation of this hypothesis to future work.

\paragraph{Implications for evaluation reporting.}
These results suggest that reporting a single $(J, f, D)$ configuration can substantially misrepresent the underlying behavior of a judge, especially on format-sensitive benchmarks. For example, Gemini-2.5-Pro appears among the strongest judges on JudgeBench under VF-JSON prompting, yet among the weakest under Direct prompting. We therefore recommend reporting either the interval
$(T_{\min}, T_{\max})$
across a fixed format set or, equivalently,
$(T_{\min}, \Delta T_{\text{fmt}})$
when comparing judge reliability across models.

\newpage
\section{Variance Decomposition Details}
\label{app:gtheory}

This appendix reports the variance decomposition referenced in
\S\ref{sec:findings-variation}. We pursue two complementary
analyses: an ANOVA on the trustworthy rate $T$ (\S\ref{app:tanova})
and a Generalizability-theory study on verdict direction
(\S\ref{app:gstudy}).

\subsection{ANOVA on $T$}
\label{app:tanova}

Treating each of the 120 (judge, format, dataset) cells at
$T_s\!=\!0$ as an observation, we fit a three-factor ANOVA with
judge, format, and dataset as factors and their pairwise
interactions. Table~\ref{tab:tanova} reports the resulting
$\eta^2$ values.

\begin{table}[h]
\centering\small
\begin{tabular}{lr}
\toprule
\textbf{Source} & \boldmath$\eta^2$ \textbf{(\%)} \\
\midrule
Judge                    & 27.0 \\
Dataset                  & 26.2 \\
Judge $\times$ Dataset   & \textbf{28.1} \\
Judge $\times$ Format    &  8.3 \\
Format                   &  0.7 \\
Format $\times$ Dataset  &  1.3 \\
Residual                 &  8.4 \\
\bottomrule
\end{tabular}
\caption{Variance decomposition ($\eta^2$, \%) of $T$ across 120
(judge, format, dataset) cells. The judge\,$\times$\,dataset
interaction is the dominant source.}
\label{tab:tanova}
\end{table}

\subsection{Four-Facet G-Study on Verdict Direction}
\label{app:gstudy}

Standard G-theory analyses for LLM judges treat the design as a
two-facet (pair\,$\times$\,judge) model and collapse format and
order into residual variance \citep{brennan2001generalizability}.
This obscures two effects that $T$ directly measures: order
sensitivity (mapped to $\varphi$) and stochastic noise (mapped
to $C$). We instead estimate a four-facet fully crossed model ---
pair ($p$) $\times$ judge ($j$) $\times$ format ($f$) $\times$
order ($o$) --- with repetitions nested within cells.
Table~\ref{tab:gtheory4} reports the seven variance components
plus residual for verdict direction at $T_s\!=\!0$. Components
shown are percentages of the total variance estimated by Restricted Maximum Likelihood;
the remainder is the pair main effect ($\sigma^2_p$).

\begin{table}[h]
\centering\small
\begin{tabular}{l rrr rrrr}
\toprule
 & \multicolumn{3}{c}{\textbf{Main effects}}
 & \multicolumn{4}{c}{\textbf{Interactions + residual}} \\
\cmidrule(lr){2-4}\cmidrule(lr){5-8}
\textbf{Dataset}
 & $\sigma^2_j$ & $\sigma^2_f$ & $\sigma^2_o$
 & $\sigma^2_{pj}$ & $\sigma^2_{pf}$ & $\sigma^2_{po}$ & Res. \\
\midrule
MT-Bench    & 0.0 & 0.0 & 2.1 & 10.3 & 1.1 & 5.0 & 4.2 \\
AlpacaFarm  & 2.1 & 0.0 & 1.8 & 17.5 & 1.4 & 3.8 & 6.0 \\
JudgeBench  & 0.9 & 0.1 & \textbf{6.2} & \textbf{34.8} & 4.4 & 6.2 & 7.9 \\
LLMBar      & 0.3 & 0.0 & 2.6 & 22.7 & 1.8 & 4.2 & 4.3 \\
\bottomrule
\end{tabular}
\caption{Four-facet variance decomposition (\% of $\sigma^2_{\text{total}}$)
for verdict direction at $T_s\!=\!0$.}
\label{tab:gtheory4}
\end{table}

Three patterns emerge. First, the order main effect is the only
non-pair main effect of non-negligible size (2--6\% across
datasets, peaking at 6.2\% on JudgeBench), confirming that
position bias is a structured measurement facet --- the variance
mapped by $\varphi$ in our formalization. Second, the
pair\,$\times$\,judge interaction is the largest within-pair
source of disagreement (10--35\%); on JudgeBench it actually
exceeds the pair main effect itself, indicating that reliability
is item-specific rather than judge-level. Third, the
pair\,$\times$\,order interaction is itself substantial
(4--6\%): position bias varies across items, exactly the
pair-level structure that $\varphi$ aggregates.

\newpage
\section{Prompt Templates}
\label{app:prompts}

All prompts share a common user message template and differ only in
the system prompt and output format instruction.
Placeholders \texttt{\{question\}}, \texttt{\{response\_a\}}, and
\texttt{\{response\_b\}} are filled per evaluation pair.
When the presentation order is BA, the contents of Response~A and
Response~B are swapped; all other text is identical.

\subsection{User Message Template (All Formats)}
\label{app:prompts:user}

\begin{mdframed}[style=logitbox]
\small
\texttt{[Question]}\\
\texttt{\{question\}}\\[4pt]
\texttt{[Response A]}\\
\texttt{\{response\_a\}}\\[4pt]
\texttt{[Response B]}\\
\texttt{\{response\_b\}}\\[4pt]
\textit{\{format-specific instruction appended here\}}
\end{mdframed}

\subsection{Pairwise Evaluation (5 Formats)}
\label{app:prompts:pairwise}

\paragraph{System Prompt (shared by all 5 pairwise formats).}

\begin{mdframed}[style=logitbox]
\small
You are an expert and impartial judge evaluating the quality of two AI
assistant responses to a user question. Your task is to determine which
response is better. Focus on helpfulness, accuracy, depth, clarity, and
relevance. Avoid any position bias --- the order in which responses appear
should not influence your judgment.
\end{mdframed}

\paragraph{Format 1: Direct (baseline).}
Single-token verdict, no rationale. Max tokens: 32.

\begin{mdframed}[style=freebox]
\small
Reply with exactly one word: A, B.
A means Response A is better. B means Response B is better.
\end{mdframed}

\paragraph{Format 2: JSON Verdict-First.}
Verdict before rationale (``decide then justify''). Max tokens: 4096.

\begin{mdframed}[style=jsonbox]
\small
Reply with a JSON object only (no markdown fences), in this exact shape:\\
\texttt{\{"winner": "A", "rationale": "your reasoning here"\}}\\
winner must be A, B.
\end{mdframed}

\paragraph{Format 3: JSON CoT (rationale-first).}
Rationale before verdict (``think then decide''). Max tokens: 4096.

\begin{mdframed}[style=jsonbox]
\small
Reply with a JSON object only (no markdown fences), in this exact shape:\\
\texttt{\{"rationale": "your reasoning here", "winner": "A"\}}\\
winner must be A, B.
\end{mdframed}

\paragraph{Format 4: XML Verdict-First.}
Verdict before rationale in XML. Max tokens: 4096.

\begin{mdframed}[style=xmlbox]
\small
Reply with an XML block only, in this exact shape:\\
\texttt{<judgment><winner>A</winner><rationale>your reasoning here</rationale></judgment>}\\
winner must be A, B.
\end{mdframed}

\paragraph{Format 5: XML CoT (rationale-first).}
Rationale before verdict in XML. Max tokens: 4096.

\begin{mdframed}[style=xmlbox]
\small
Reply with an XML block only, in this exact shape:\\
\texttt{<judgment><rationale>your reasoning here</rationale><winner>A</winner></judgment>}\\
winner must be A, B.
\end{mdframed}

\subsection{GEval Rubric-Based Evaluation}
\label{app:prompts:geval}

GEval uses a single API call per evaluation with a structured rubric in the system prompt and a JSON output format requiring both scores and a verdict.

\paragraph{Shared Output Format Instruction (prepended to user message).}

\begin{mdframed}[style=jsonbox]
\small
IMPORTANT: You must respond with ONLY a JSON object. Do not write any free text, do not solve the problem yourself, do not use markdown fences. Output exactly this shape and nothing else:\\
\texttt{\{"rationale": "your evaluation here", "score\_a": 4, "score\_b": 2, "winner": "A"\}}\\
score\_a and score\_b are integers from 1 (very poor) to 5 (excellent) rating each response against the rubric criteria independently.\\
winner must be exactly one of: A, B.\\
winner should be A if score\_a > score\_b, B if score\_b > score\_a.
\end{mdframed}

\paragraph{GEval System Prompt — JudgeBench.}

\begin{mdframed}[style=loosebox]
\small
You are an expert and impartial judge evaluating the quality of two AI assistant responses to a user question. Your task is to determine which response is better based on a structured rubric.

Apply the following evaluation rubric in order. Consider criteria earlier in the list more important when responses are otherwise close.

\textbf{RUBRIC:}
\begin{enumerate}
\item \textbf{Factual Accuracy} --- Does the response contain correct facts, calculations, code, or logical deductions? Errors in reasoning, arithmetic, or factual claims are heavily penalized.
\item \textbf{Correctness of Final Answer} --- For questions with a definitive correct answer (math, science, code, multiple-choice), does the response arrive at the right answer? A response with the correct final answer but imperfect explanation is preferred over a response with an incorrect answer and polished presentation.
\item \textbf{Completeness} --- Does the response address all parts of the question without omitting required steps, sub-problems, or output requirements stated in the prompt?
\item \textbf{Clarity and Presentation} --- Is the response well-organized, clearly explained, and easy to follow? Prefer structured, step-by-step exposition where appropriate.
\end{enumerate}

\textit{Scoring guidance:}
If one response has a factual error or wrong answer and the other does not, the response without the error wins regardless of other criteria. A tie is appropriate only when both responses are correct and equivalent in quality across all criteria, or both contain equivalent errors. Avoid any position bias --- the order in which responses appear should not influence your judgment.
\end{mdframed}

\paragraph{GEval System Prompt — LLMBar.}

\begin{mdframed}[style=loosebox]
\small
You are an expert and impartial judge evaluating the quality of two AI assistant responses to a user instruction. Your task is to determine which response is better based on a structured rubric.

Apply the following evaluation rubric in order. Consider criteria earlier in the list more important when responses are otherwise close.

\textbf{RUBRIC:}
\begin{enumerate} 
\item \textbf{Instruction Adherence} --- Does the response follow every explicit and implicit constraint specified in the instruction? Violations of format, scope, perspective, tone, length, or output type are penalized. A response that is fluent but ignores or modifies the instruction is worse than a plainer response that follows it exactly.
\item \textbf{Completeness} --- Does the response address all aspects of the instruction without adding unsolicited content or dropping required elements?
\item \textbf{Relevance} --- Is the response directly focused on the task without introducing irrelevant information or tangential elaborations that were not requested?
\item \textbf{Clarity} --- Is the response well-written and easy to understand, given that it correctly addresses the instruction?
\end{enumerate}

\textit{Scoring guidance:}
Instruction adherence is the primary criterion. A response that deviates from the instruction in any material way loses to one that follows it, even if the deviating response is more informative or better written. A tie is appropriate only when both responses follow the instruction equally and are equivalent in all other criteria. Avoid any position bias --- the order in which responses appear should not influence your judgment.
\end{mdframed}

\subsection{FLASK Multi-Criterion Evaluation}
\label{app:prompts:flask}

FLASK issues 4 separate API calls per evaluation one per criterion
and aggregates by majority vote.
Each call uses the same JSON output format as GEval
(Section~\ref{app:prompts:geval}) with an additional criterion
definition injected into the user message.

\paragraph{FLASK System Prompt — JudgeBench.}

\begin{mdframed}[style=freebox]
\small
You are an expert and impartial judge evaluating the quality of two AI assistant responses to a user question. Your task is to determine which response is better on one specific criterion. Evaluate only the criterion stated --- do not factor in other quality dimensions. Avoid any position bias.
\end{mdframed}

\paragraph{FLASK System Prompt — LLMBar.}

\begin{mdframed}[style=freebox]
\small
You are an expert and impartial judge evaluating the quality of two AI assistant responses to a user instruction. Your task is to determine which response is better on one specific criterion. Evaluate only the criterion stated --- do not factor in other quality dimensions. Avoid any position bias.
\end{mdframed}

\paragraph{FLASK Criteria — JudgeBench.}
The following criterion definitions are injected into the user message as
\texttt{[Criterion: \{name\}]} blocks.
Each is evaluated independently in a separate API call.

\begin{mdframed}[style=loosebox]
\small
\textbf{1. Factual Accuracy.}
Which response contains fewer factual errors, incorrect calculations, or flawed reasoning steps? A response with any factual error loses to one that is entirely correct.

\textbf{2. Completeness.}
Which response more fully addresses all parts of the question, including all required sub-problems, output formats, or conditions stated in the prompt?

\textbf{3. Logical Coherence.}
Which response presents a more logically consistent chain of reasoning, with conclusions that follow from stated premises and no internal contradictions?

\textbf{4. Instruction Following.}
Which response more precisely follows the structural and formatting requirements stated in the question (e.g., box the answer, return JSON, produce code in a specific language)?
\end{mdframed}

\paragraph{FLASK Criteria — LLMBar.}

\begin{mdframed}[style=loosebox]
\small
\textbf{1. Instruction Adherence.}
Which response more faithfully follows every explicit and implicit constraint in the instruction --- including required format, scope, perspective, tone, length, and output type --- without omitting or modifying any stated requirement?

\textbf{2. Relevance.}
Which response stays more tightly focused on the requested task, without adding unsolicited content, tangential elaborations, or off-topic information?

\textbf{3. Completeness.}
Which response more fully addresses all elements of the instruction without omitting required parts or prematurely terminating the response?

\textbf{4. Clarity.}
Which response is better written, more clearly organized, and easier to understand, assuming it has otherwise satisfied the instruction?
\end{mdframed}

\subsection{Prompt Design Summary}
\label{app:prompts:summary}

\begin{table}[h]
\centering
\small
\begin{tabular}{llcccl}
\toprule
\textbf{Framework} & \textbf{Format} & \textbf{Rubric} & \textbf{Scores} & \textbf{Cost} & \textbf{Output} \\
\midrule
Pairwise & Direct       & No  & No  & 1 & A / B\\
Pairwise & JSON VF      & No  & No  & 1 & JSON \\
Pairwise & JSON CoT     & No  & No  & 1 & JSON \\
Pairwise & XML VF       & No  & No  & 1 & XML \\
Pairwise & XML CoT      & No  & No  & 1 & XML \\
\cmidrule(l){1-6}
GEval    & JSON CoT     & Yes & 1--5 & 1 & JSON \\
\cmidrule(l){1-6}
FLASK    & JSON CoT     & Yes & 1--5 & 4 & JSON $\times$ 4 \\
\bottomrule
\end{tabular}
\caption{Prompt design matrix.
All frameworks share the same user message template (Section~\ref{app:prompts:user}).
Cost is measured in API calls per (pair, judge, order) evaluation.}
\label{tab:prompt_summary}

\end{table}

\section{API Cost}
\label{app:cost}

The total API cost for conducting the cross-cutting reliability audit was approximately \$2,100 USD. A detailed breakdown of costs across the three evaluation paradigms is provided below (based on public API cost of model providers):

\begin{table}[h]
\centering
\begin{tabular}{lr}
\hline
\textbf{Evaluation Paradigm} & \textbf{Cost (USD)} \\ \hline
Pairwise Experiments        & \$1,450             \\
FLASK Rubric Experiments     & \$390               \\
G-Eval Experiments           & \$260               \\ \hline
\textbf{Total}               & \textbf{\$2,100}    \\ \hline
\end{tabular}
\caption{Breakdown of API costs across experimental paradigms.}
\label{tab:api_costs}
\end{table}

\newpage
\section{Dataset Details}
\label{app:datasets}

Table~\ref{tab:datasets} summarizes the four benchmarks used in our study. All pairs were fixed at the start of the experiment and shared across all judges, formats, presentation orders, temperatures, and repetitions.

\begin{table*}[h!]
\centering
\small
\begin{tabular}{llccl}
\toprule
\textbf{Dataset} & \textbf{Truth} & \textbf{Pairs}
& \textbf{Domains} & \textbf{Sampling} \\
\midrule
MT-Bench
  & Subjective & 100 & 1
  & Random (seed=42) \\
AlpacaFarm
  & Subjective & 100 & 1
  & Random (seed=42) \\
JudgeBench
  & Objective & 100 & 17
  & Stratified by domain \\
LLMBar
  & Objective & 100 & 5
  & Stratified by split \\
\bottomrule
\end{tabular}
\caption{Dataset summary. Calls are computed at $T_s{=}0$ as pairs $\times$ 6 judges $\times$ 5 formats $\times$ 2 orders $\times$ 10 reps.}

\label{tab:datasets}
\end{table*}

\paragraph{MT-Bench} \cite{zheng2023judging}
Pairwise comparisons from the human judgment set. Labels are crowd-sourced preferences.

\paragraph{AlpacaFarm} \cite{dubois2023alpacafarm}
100 instruction-following pairs with binary preference labels. The balanced label distribution makes the dataset sensitive to position bias.

\paragraph{JudgeBench} \cite{tan2025judgebench}
100 pairs stratified from 620 examples across 17 domains with objective ground truth. One response in each pair is verifiably correct.

\paragraph{LLMBar} \cite{zeng2024evaluating}
100 pairs sampled evenly across 5 robustness splits. Most pairs contain adversarial distractors designed to fool LLM judges.


\section{Assumptions underlying $T$}
\label{app:a2_clarification}

The elegant generalized expression for the trustworthy verdict rate,
$T = C \cdot (A - \varphi q),$
relies on Assumption 2 (A2): that is $N$, $P$, and $R$ are mutually independent. In practice, this assumption may not hold exactly. Here we empirically demonstrate that even when A2 is violated, $T$ continues to behave as a conservative lower bound.

To test this, we define $T_{\text{joint}} = \Pr(N=1 \wedge P=1 \wedge R=1)$, the assumption-free joint probability, and compare it against the closed-form estimate $T_{\text{closed}} = C \cdot (A - \varphi/2)$.

Table \ref{tab:benchmark_comparison} reports the gap $T_{\text{closed}} - T_{\text{joint}}$ across benchmarks and judges. The gap is consistently small (typically under 5 pp.), indicating that the closed-form expression remains an empirically bounded, conservative approximation even without strict independence.

\begin{table}[h!]
\centering
\small
\begin{tabular}{llcccc}
\toprule
\textbf{Benchmark} & \textbf{Judge} & \textbf{$T_{\text{closed}}$} & \textbf{$T_{\text{joint}}$} & \textbf{Gap} & \textbf{Gap 95\% CI} \\
\midrule
\multirow{6}{*}{MT-Bench}
 & GPT-5.2            & 0.54 & 0.58 & -0.04 & [-0.068, -0.012] \\
 & GPT-4o             & 0.55 & 0.57 & -0.02 & [-0.041, -0.007] \\
 & Gemini-2.5-Pro     & 0.50 & 0.55 & -0.05 & [-0.072, -0.024] \\
 & Gemini-2.0-Flash   & 0.51 & 0.52 & -0.01 & [-0.013, -0.001] \\
 & Claude-Sonnet-4.6  & 0.62 & 0.64 & -0.02 & [-0.043, -0.003] \\
 & Llama-3.3-70B      & 0.58 & 0.58 & 0.00  & [-0.006, +0.000] \\
\midrule
\multirow{6}{*}{JudgeBench}
 & GPT-5.2            & 0.49 & 0.55 & -0.05 & [-0.082, -0.023] \\
 & GPT-4o             & 0.31 & 0.32 & 0.00  & [-0.012, +0.023] \\
 & Gemini-2.5-Pro     & 0.62 & 0.65 & -0.03 & [-0.063, +0.000] \\
 & Gemini-2.0-Flash   & 0.23 & 0.23 & 0.00  & [-0.008, +0.008] \\
 & Claude-Sonnet-4.6  & 0.64 & 0.67 & -0.03 & [-0.058, -0.010] \\
 & Llama-3.3-70B      & 0.32 & 0.32 & 0.00  & [-0.001, +0.005] \\
\midrule
\multirow{6}{*}{LLMBar}
 & GPT-5.2            & 0.67 & 0.70 & -0.04 & [-0.062, -0.015] \\
 & GPT-4o             & 0.64 & 0.66 & -0.02 & [-0.032, -0.002] \\
 & Gemini-2.5-Pro     & 0.65 & 0.68 & -0.03 & [-0.053, -0.008] \\
 & Gemini-2.0-Flash   & 0.43 & 0.43 & 0.00  & [-0.012, +0.008] \\
 & Claude-Sonnet-4.6  & 0.76 & 0.77 & -0.01 & [-0.021, +0.001] \\
 & Llama-3.3-70B      & 0.58 & 0.58 & 0.00  & [-0.005, +0.000] \\
\midrule
\multirow{6}{*}{AlpacaFarm}
 & GPT-5.2            & 0.40 & 0.43 & -0.03 & [-0.058, +0.002] \\
 & GPT-4o             & 0.48 & 0.50 & -0.02 & [-0.032, +0.001] \\
 & Gemini-2.5-Pro     & 0.39 & 0.41 & -0.02 & [-0.045, +0.005] \\
 & Gemini-2.0-Flash   & 0.43 & 0.43 & 0.00  & [-0.007, +0.008] \\
 & Claude-Sonnet-4.6  & 0.51 & 0.53 & -0.02 & [-0.033, +0.000] \\
 & Llama-3.3-70B      & 0.52 & 0.52 & 0.00  & [-0.005, +0.000] \\
\bottomrule
\end{tabular}
\caption{Comparison of $T_{\text{closed}}$ and $T_{\text{Joint}}$ across benchmarks and judges}
\label{tab:benchmark_comparison}

\end{table}

\end{document}